\documentclass{article} 
\PassOptionsToPackage{table}{xcolor}

\usepackage{amsmath,amsfonts,bm}

\renewcommand{\eqref}[1]{\textup{(\ref{#1})}}

\def\1{\bm{1}}

\DeclareMathAlphabet{\mathsfit}{\encodingdefault}{\sfdefault}{m}{sl}
\SetMathAlphabet{\mathsfit}{bold}{\encodingdefault}{\sfdefault}{bx}{n}

 \usepackage{RayCo4D}
\usepackage{hyperref}
\usepackage{url}
\usepackage{graphicx}
\usepackage[table]{xcolor}
\usepackage{float}
\usepackage{multirow}
\usepackage{booktabs}
\usepackage{array}
\usepackage{tabularx}
\usepackage{ltablex}
\usepackage{wrapfig}
\usepackage{needspace}
\usepackage{capt-of}

\title{RayCo4D: Ray-Conditioned Continuous Motion Model for Monocular 4D Dynamic Reconstruction}

\orglabel{}
\orglogo{%
  \makebox[\linewidth][l]{%
    \includegraphics[height=1.2cm,keepaspectratio]{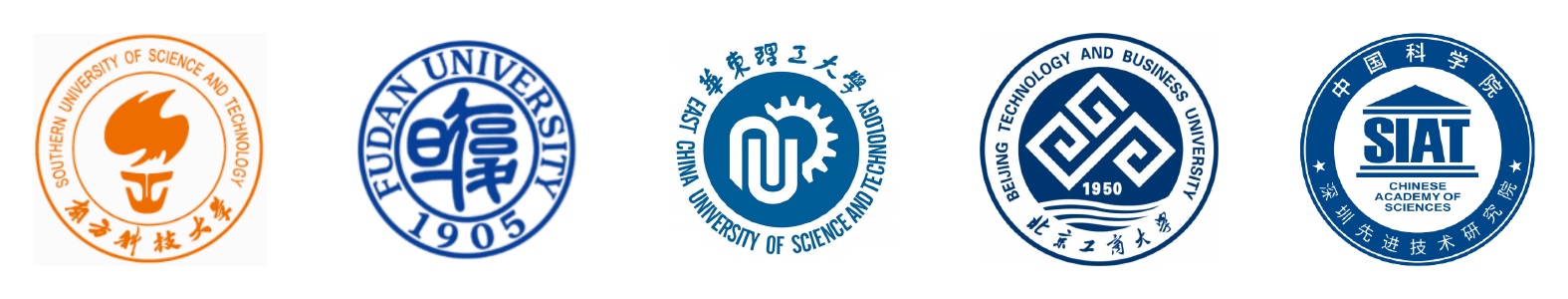}%
  }%
}

\contributors{%
  \small
  Yuanxiang Ni\textsuperscript{1}\quad
  Xianliang Huang\textsuperscript{2,\dag}\quad
  Ruihan Hou\textsuperscript{3}\quad Feng Xie\textsuperscript{4}\quad 
  Hao Zhang\textsuperscript{5,*}\quad
}

\contriblegend{
    \textsuperscript{1} Southern University of Science and Technology, Shenzhen, China \\
    \textsuperscript{2} Fudan University, Shanghai, China \\
    \textsuperscript{3} East China University of Science and Technology, Shanghai, China \\ 
    \textsuperscript{4} Beijing Technology and Business University, Beijing, China \\ \textsuperscript{5} Shenzhen Institutes of Advanced Technology, Chinese Academy of Sciences, Shenzhen, China \\
    \textsuperscript{*}\,Corresponding author: \href{mailto:h.zhang10@siat.ac.cn}{\texttt{h.zhang10@siat.ac.cn}} \\
    \textsuperscript{\dag}\,Project leader: \href{mailto:huangxl21@m.fudan.edu.cn}{\texttt{huangxl21@m.fudan.edu.cn}}
}

\projectlinks{%
  \link{Project Page}{https://123n123.github.io/rayco4d-demo/}\quad
  \link{Code}{https://github.com/123n123/RayCo4D}
  }

\begin{document}
\maketitle

\begin{figure}[htbp]
    \centering
    \includegraphics[width=1\linewidth]{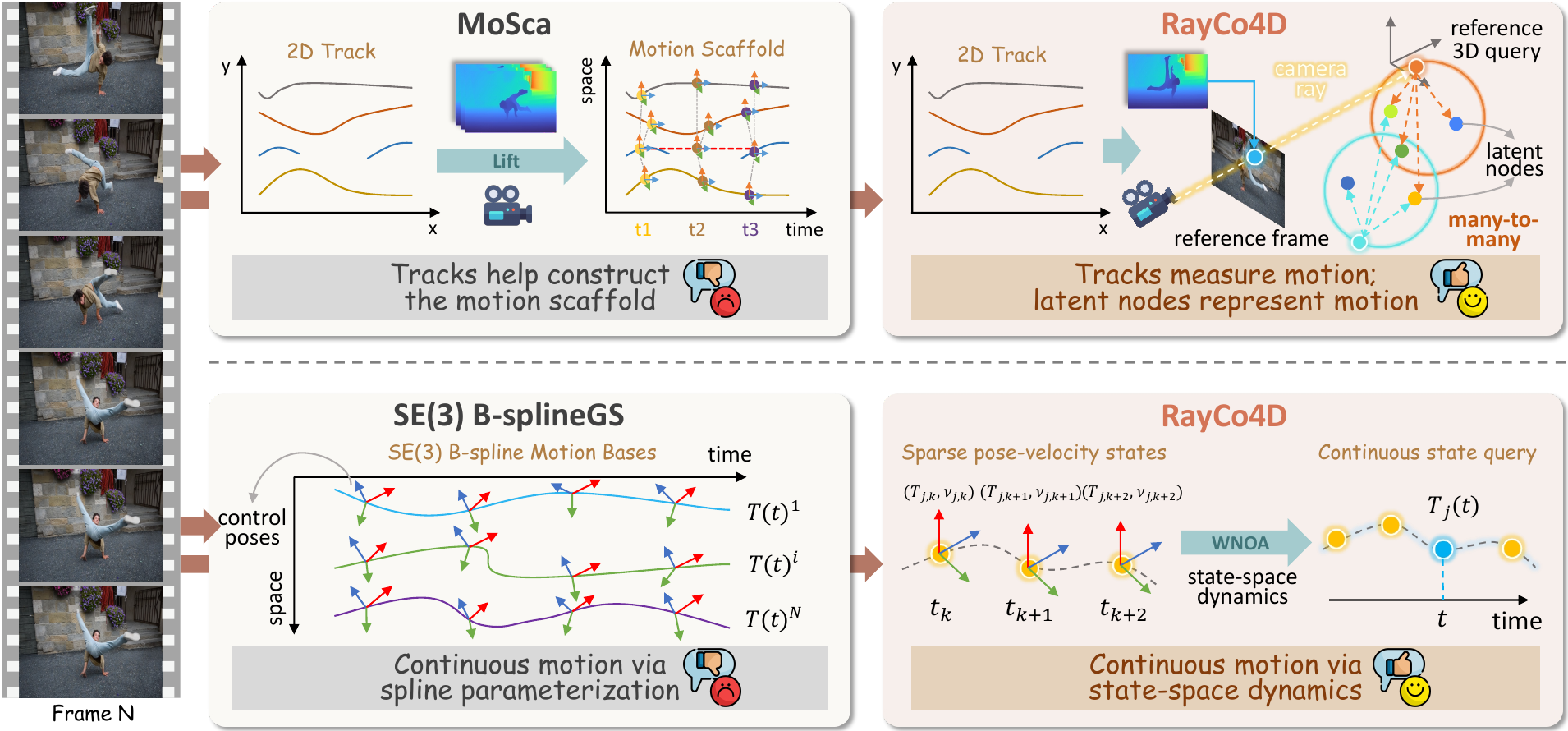}
    \vspace{-0.3cm}
    \caption{\textbf{RayCo4D versus prior track-driven methods.} \textbf{Top:} MoSca~\citep{lei2025mosca} initializes motion support from lifted tracks; RayCo4D treats tracks as ray-conditioned measurements of shared latent nodes, without one-to-one track--node binding. \textbf{Bottom:} SE(3) B-splineGS~\citep{zhang2026se3bspline} utilizes continuous motion with spline trajectories, whereas RayCo4D models it using sparse $\mathrm{SE}(3)$ pose-velocity states with a WNOA prior for continuous-time queries.}
    \label{fig:teaser}
\end{figure}

\begin{abstract}
Reconstructing dynamic 3D scenes from monocular video requires inferring motion from noisy and intermittently visible observations. Long-range tracks provide valuable geometric constraints, but integrating their fragmented temporal coverage into shared, continuous motion remains challenging.
We introduce \textbf{RayCo4D}, a ray-conditioned continuous motion model for monocular 4D reconstruction that explicitly separates visual measurements from latent 3D motion states.
Each track is associated with a reference-time 3D query constrained to its camera ray, and its image and depth observations constrain shared latent motion nodes through local many-to-many associations. Each node maintains an $\mathrm{SE}(3)$ pose and body-frame velocity at sparse support times. A white-noise-on-acceleration process prior couples neighboring states and defines continuous-time interpolation, allowing observations across visibility intervals to jointly constrain the same latent state sequence.
The inferred shared motion is transferred to Gaussian primitives through dual-quaternion blending, while reference-anchored local $\mathrm{SE}(3)$ residuals capture fine-scale non-rigid deformation beyond the shared scaffold. Experiments on diverse monocular dynamic scenes show that RayCo4D achieves competitive novel-view synthesis quality, improves structural and temporal consistency under rapid motion and incomplete tracks, and supports continuous-time rendering.

\end{abstract}

\section{Introduction}
\label{sec:introduction}
Reconstructing 3D scenes from images is an important step toward persistent and renderable representations of the physical world~\citep{guo2026salon3r,shu2026litevggt,zhou2026think}. Such representations can support applications, such as 3D content creation~\citep{11731530,wu2025cat4d,yan2025instant,ni2026cogeo}, robot perception~\citep{song2025robospatial, jia2025lift3d, huang2026semantic}, and dynamic scene understanding~\citep{lei2025mosca,wang2025shapeofmotion,zhu2026mint}.
In monocular dynamic reconstruction, only one viewpoint is available at each instant, leaving scene geometry, camera motion, and non-rigid deformation jointly under-constrained. Occlusions, motion blur, and tracking failures further fragment the cross-frame evidence available for motion estimation.

Recent methods build on the explicit representation and efficient rendering of 3D Gaussian Splatting~\citep{kerbl2023gaussians}, using long-range tracks~\citep{karaev2024cotracker} and monocular depth to constrain dynamic scene reconstruction.
An important modeling choice is how these geometric cues relate to the underlying motion states.
MoSca~\citep{lei2025mosca} initializes motion-node trajectories from lifted tracks, while SE(3) B-splineGS~\citep{zhang2026se3bspline} initializes continuous spline control poses from lifted tracklets. Shape of Motion~\citep{wang2025shapeofmotion} uses tracks to supervise shared $\mathrm{SE}(3)$ motion bases, illustrating the value of pooling observations through a common motion representation. For incomplete tracks, the central challenge is to combine measurements
across visibility intervals into locally shared motion states with coherent continuous-time evolution.

Our central perspective is that \emph{image tracks are measurements of latent 3D motion, rather than the motion states themselves}.
Based on this perspective, we introduce \textbf{RayCo4D}, a ray-conditioned continuous motion model for monocular 4D reconstruction. Fig.~\ref{fig:teaser} summarizes its two central design choices:
ray-conditioned measurement--state modeling and continuous pose--velocity state estimation.
Using camera and depth estimates from Depth Anything 3~\citep{lin2025depth}, we constrain each track with a reference-time 3D query and use its visible observations to estimate the shared motion.
Track geometry helps initialize a compact motion support, but subsequent optimization does not maintain a dedicated motion state for each track.
Instead, each query is associated with a local neighborhood of motion nodes. A track can constrain multiple nodes, while each node aggregates observations from multiple tracks and timestamps.
This many-to-many association allows track observations with different visibility intervals to contribute to a shared motion representation.

To estimate this shared motion continuously, each node maintains an $\mathrm{SE}(3)$ pose and body-frame velocity at sparse support times.
We connect consecutive states using a white-noise-on-acceleration (WNOA) process prior from continuous-time state estimation~\citep{barfoot2024state}.
The process prior couples neighboring pose--velocity states and defines consistent interpolation between support times.
Track and depth measurements constrain node transformations at their observation times.
As a result, observations from different frames and visibility intervals contribute to the same latent state sequence.
Intervals without direct observations remain connected through the process model. Temporal continuity enters both the motion representation and its estimation.

To obtain renderable dynamic geometry, we propagate the node transformations to Gaussian primitives through dual-quaternion blending.
This shared scaffold provides a common motion structure, but cannot capture every fine-scale non-rigid deformation.
We therefore augment each dynamic Gaussian with a local $\mathrm{SE}(3)$ residual expressed in its reference configuration.
The residual transform is exactly the identity at the Gaussian's reference time, ensuring that it introduces no local correction at that anchor.
Together with regularization on residual magnitude and variation, this design encourages local corrections to complement, rather than replace, the shared motion. RayCo4D thus combines continuous shared motion with reference-anchored local refinement in a hierarchical dynamic representation. 

Our main contributions are summarized as follows: (1) We introduce a \textbf{ray-conditioned measurement--state formulation} in which 2D tracks and camera--depth geometry constrain shared latent motion nodes through many-to-many associations, decoupling track identity from motion-state identity. 
(2) We formulate shared motion using sparse continuous-time
$\mathrm{SE}(3)$ pose-velocity states, with WNOA process factors coupling neighboring supports and defining interpolation for measurement evaluation and rendering.
(3) We develop a \textbf{hierarchical Gaussian motion representation} that combines shared node transport with reference-anchored local $\mathrm{SE}(3)$ residuals, capturing coherent structural motion and fine-scale non-rigid deformation. 
(4) Experiments demonstrate competitive novel-view synthesis and improved structural and temporal consistency under fast motion and incomplete tracks, together with continuous-time rendering within the reconstructed interval.

\section{Related Work}
\label{sec:related_work}
\paragraph{Structured Motion for Dynamic Gaussian Splatting.}
Time-conditioned Gaussian deformation is used to model dynamic scenes~\citep{yang2024deformable3dgs,wu2024fourDgs,xu2024grid4d}. Structured representations introduce shared motion across primitives: Shape of Motion~\citep{wang2025shapeofmotion} fits shared $\mathrm{SE}(3)$ motion bases using track and depth supervision, while MoSca~\citep{lei2025mosca} initializes sparse motion nodes from lifted tracks and propagates their transformations to Gaussians.
OriGS~\citep{wu2025origs}, RiGS~\citep{wu2026rigs}, and USPLAT4D~\citep{guo2026usplat4d} further introduce orientation fields, motion-type decomposition, and uncertainty-guided motion graphs,
respectively. Building on shared motion representations, RayCo4D connects ray-conditioned track measurements to latent nodes through local many-to-many associations and augments their motion with
reference-anchored Gaussian-level corrections.

\paragraph{Continuous Motion and State-Space Modeling.}
SplineGS~\citep{park2025splinegs} represents Gaussian positions with cubic Hermite splines, while
SE(3) B-splineGS~\citep{zhang2026se3bspline} models position and orientation using cumulative $\mathrm{SE}(3)$ B-splines initialized from lifted tracklets.
Spline-based motion models are also used for dynamic scene reconstruction in autonomous driving~\citep{xu2025adgs}.
GP-4DGS~\citep{kim2026gp4dgs} uses variational Gaussian processes to model correlated deformations and their uncertainty. Related dynamical formulations include prediction--observation fusion
over Gaussian states in Gaussians on their Way~\citep{deng2025gaussians}, learned velocity fields in EvoGS~\citep{asiimwe2025evogs}, and latent ODE dynamics in ODE-GS~\citep{wang2026odegs}.
RayCo4D adopts a WNOA prior from continuous-time state estimation~\citep{barfoot2014continuous,barfoot2024state} to jointly estimate shared node poses and body velocities at sparse support times. The same pose-velocity model defines temporal process factors and continuous motion queries, allowing ray-based observations to constrain a common latent state sequence. Additional discussion is provided in Appendix~\ref{sec:additional_related_work}.

\section{Method}
\label{sec:method}

\begin{figure*}[t]
    \centering
    \includegraphics[width=\textwidth]{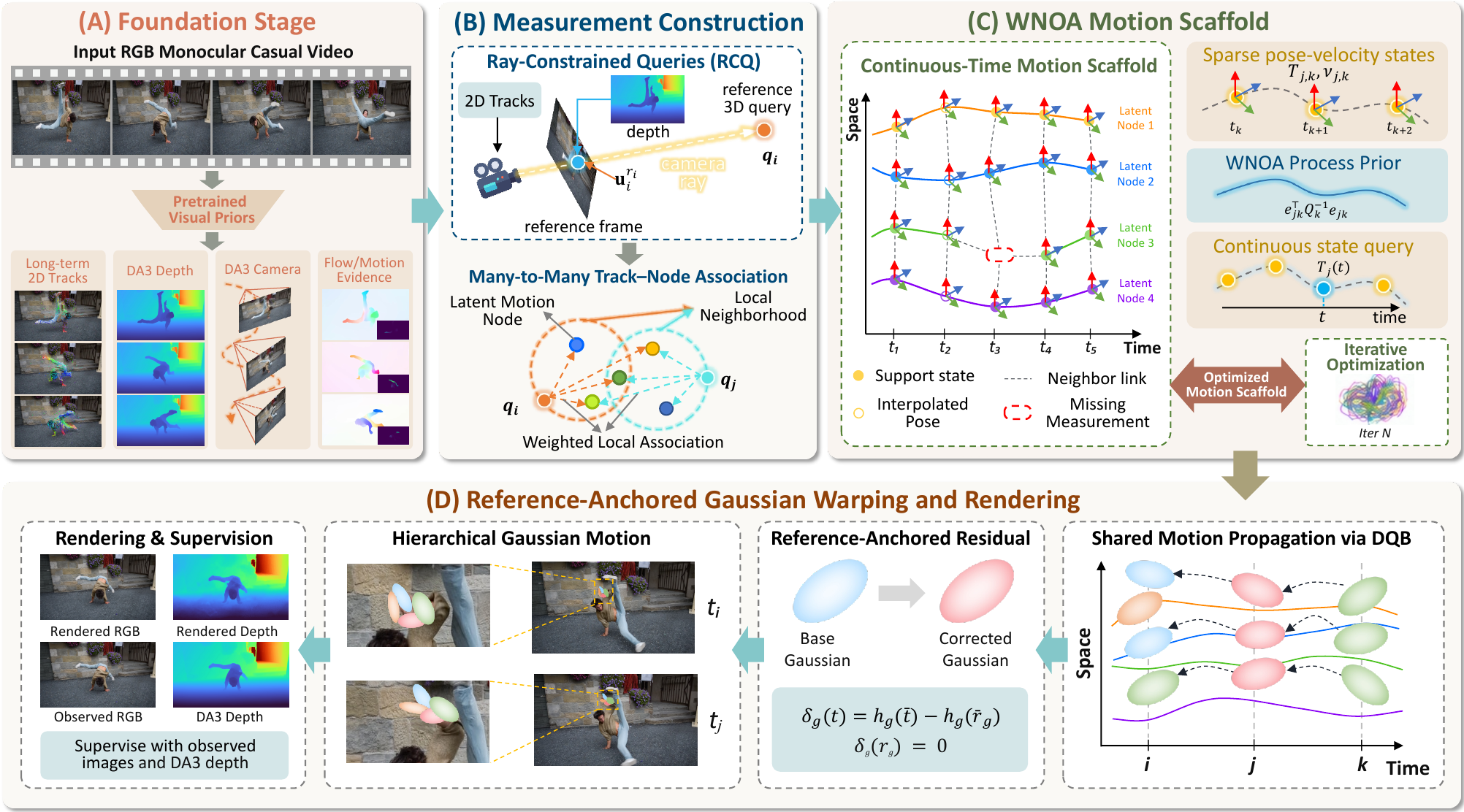}
    \vspace{-0.3cm}
    \caption{\textbf{Overview of RayCo4D.} Given a monocular video, RayCo4D constructs ray-conditioned 3D queries from 2D tracks and camera-depth estimates to constrain shared latent motion nodes. Sparse SE(3) pose--velocity states are coupled by a WNOA prior to model continuous motion. The inferred motion is propagated to Gaussians via dual-quaternion blending, while reference-anchored local residuals capture fine-grained non-rigid deformation for dynamic rendering and optimization.}
    \label{fig:system_overview}
\end{figure*}

Given a monocular video $\{I_f\}_{f=0}^{F-1}$, RayCo4D reconstructs a dynamic Gaussian scene by separating visual measurements from shared continuous motion states (Fig.~\ref{fig:system_overview}). The inputs comprise 2D tracks with visibility and confidence, camera and depth estimates, and epipolar cues derived from RAFT~\citep{teed2020raft} optical flow. Our representation consists of a sparse set of latent motion nodes, reference-time Gaussian geometry, and per-Gaussian local motion corrections. Through ray-conditioned queries, tracks constrain the motion nodes, whose transformations also drive the Gaussian scene. We formulate ray-conditioned measurements, continuous pose–velocity dynamics, and reference-anchored local corrections.

\subsection{Ray-Conditioned Motion Measurements}
\label{sec:ray_measurements}
Long-range tracks connect observations across time, but their visibility, confidence, and depth support vary across frames. The goal is to convert these fragmented observations into geometric constraints on a compact, shared motion field.
RayCo4D anchors each track to a camera ray in a reliable reference frame and associates its 3D query with nearby latent motion nodes. The resulting node-induced motion is then evaluated against image and depth measurements.
\paragraph{Ray-constrained queries (RCQ).}
To couple a 2D track with the 3D motion field, we first assign it a reference location in world space. For each track $i$, let $\mathbf{u}_{i,f}$ denote its observed
pixel coordinate at frame $f$, and $r_i$ a visible reference frame
selected based on observation reliability.
We represent the reference-time track geometry as a 3D query
$\mathbf{q}_i$, whose position is constrained and optimized
along the corresponding camera ray:
\begin{equation}
\begin{aligned}
 z_i
 &=
 z_{\min}+(d_i^0-z_{\min})\exp(\eta_i), \quad \quad
 \mathbf q_i
 &=
 \mathbf c_{r_i}
 +z_i\mathbf R_{r_i}^c\mathbf K_{r_i}^{-1}
 \widetilde{\mathbf u}_{i r_i},
\end{aligned}
\label{eq:ray_query}
\end{equation}
where $\widetilde{\mathbf u}$ denotes homogeneous pixel coordinates, $z_{\min}>0$ is a depth floor, and $\eta_i$ is an optimizable scalar initialized to zero. Here, $\mathbf c_{r_i}$, $\mathbf R_{r_i}^c$, and $\mathbf K_{r_i}$ denote the world-space camera center, camera-to-world rotation, and intrinsics at frame $r_i$, respectively.
The initial depth $d_i^0$ is taken from a reliable depth observation. For tracks without valid depth, it is initialized from a front-facing motion node whose projection is close to the reference pixel.
The unnormalized ray in Eq.~\eqref{eq:ray_query} preserves the camera-axis depth convention.
Importantly, $\eta_i$ adjusts reference geometry and it does not parameterize time-varying track motion.

\paragraph{Measurement--state decoupling (MSD).}
Let $\mathbf T_j(t)=[\mathbf R_j(t),\mathbf p_j(t)]\in\mathrm{SE}(3)$ denote the continuous pose of latent node $j$. For each query, we select its $K$ nearest nodes at the reference time and compute normalized radial weights:
\begin{equation}
\begin{aligned}
 \mathcal N_i
 &=
 \operatorname{KNN}_j
 \bigl(\mathbf q_i,\mathbf p_j(r_i)\bigr), \quad \quad
 a_{ij}
 &=
 \frac{
 \exp\!\left(
 -\|\mathbf q_i-\mathbf p_j(r_i)\|_2^2/(2\sigma_j^2)
 \right)
 }{
 \sum_{\ell\in\mathcal N_i}
 \exp\!\left(
 -\|\mathbf q_i-\mathbf p_\ell(r_i)\|_2^2/(2\sigma_\ell^2)
 \right)
 },
 \qquad j\in\mathcal N_i,
\end{aligned}
\label{eq:ray_association}
\end{equation}
where $\sigma_j$ controls the spatial influence of node $j$. Candidate neighborhoods are periodically refreshed as geometry changes, while the weights are recomputed from the current query and node positions. Queries in the same moving region consequently have overlapping node
support, allowing their observations to jointly shape the local motion.

We combine the relative transformations of these nodes using dual-quaternion blending (DQB)~\citep{kavan2007skinning}. The predicted world-space position of track $i$ at time $t$ is:
\begin{equation}
 \widehat{\mathbf x}_i(t)
 =
 \operatorname{DQB}\!\left(
 \left\{
 a_{ij},
 \mathbf T_j(t)\mathbf T_j(r_i)^{-1}
 \right\}_{j\in\mathcal N_i}
 \right)
 \cdot\mathbf q_i,
\label{eq:track_prediction}
\end{equation}
where $\cdot$ denotes the rigid action on a 3D point. Each observation constrains a local blend of node states, allowing shared nodes to aggregate evidence across tracks and time. This decouples track identity from latent motion-state identity.

\paragraph{Measurement factors.}
Let $\mathcal O_u$ denote visible track observations and $\mathcal O_d\subseteq\mathcal O_u$ those with valid depth. Keeping the observed coordinates, visibility, and confidence fixed, we constrain the predicted query positions with the confidence-weighted reprojection loss:
\begin{equation}
 \mathcal L_{\mathrm{2D}}
 =
 \frac{1}{Z_u}
 \sum_{(i,f)\in\mathcal O_u}
 c_{if}\,
 \rho_2\!\left(
 \frac{\pi_f(\widehat{\mathbf x}_i(f))-\mathbf u_{if}}
 {\sigma_u}
 \right),
 \qquad
 Z_u=\sum_{(i,f)\in\mathcal O_u}c_{if}.
\label{eq:track_loss}
\end{equation}
Here $c_{if}$ is track confidence, $\pi_f$ is camera projection, $\sigma_u$ is the image-space noise scale, and $\rho_2$ averages a componentwise smooth-$\ell_1$ penalty. Observations in $\mathcal O_d$ also contribute a relative camera-axis depth loss weighted by track confidence and depth reliability, denoted $\mathcal L_{\mathrm{depth}}$; visible observations without depth still contribute to Eq.~\eqref{eq:track_loss}.

\subsection{Continuous-Time Latent Motion with WNOA}
\label{sec:continuous_motion}
The measurement factors above reveal motion at a sparse set of observation times. The latent motion representation must ensure temporal coherence and provide node transformations at arbitrary times.
RayCo4D therefore represents each node with sparse $\mathrm{SE}(3)$ pose--velocity states and connects adjacent supports using a white-noise-on-acceleration (WNOA) process. The same continuous model is used for measurement evaluation, Gaussian motion, and subframe rendering.

\paragraph{Sparse pose--velocity states.}
At each support time, we represent each node by its pose and body-frame velocity to capture both configuration and local motion. For support times $0=\tau_0<\cdots<\tau_{S-1}=F-1$, we define:
\begin{equation}
 \mathcal X
 =
 \left\{
 \mathbf s_{jk}
 =
 (\mathbf T_{jk},\boldsymbol\nu_{jk})
 \right\}_{j,k},
 \qquad
 \mathbf T_{jk}\in\mathrm{SE}(3),
 \quad
 \boldsymbol\nu_{jk}\in\mathbb R^6.
\label{eq:motion_states}
\end{equation}
where $\mathbf T_{jk}=\mathbf T_j(\tau_k)$ and $\boldsymbol\nu_{jk}$ is the body-frame velocity.
We order the translational components before the angular components and use the convention $\dot{\mathbf T}_j(t) =\mathbf T_j(t)\boldsymbol\nu_j(t)^\wedge$.
The support times are sampled at a fixed frame stride and include both sequence endpoints.
Both pose and velocity are jointly optimized using visual measurements.

\paragraph{WNOA process model.}
The sparse states require a temporal relation that determines how motion evolves between consecutive supports. We impose a locally linear WNOA prior on $\mathrm{SE}(3)$~\citep{anderson2015full}. For the interval $[\tau_k,\tau_{k+1}]$, let
$
\Delta_k=\tau_{k+1}-\tau_k,
\boldsymbol\xi_{jk}^{+}
=
\operatorname{Log}
\bigl(\mathbf T_{jk}^{-1}\mathbf T_{j,k+1}\bigr)^\vee
$
denote the relative endpoint pose in the tangent space centered at
$\mathbf T_{jk}$.
Because the endpoint body velocities live in different local frames, we
express the terminal velocity in the same tangent space using
$
\widetilde{\boldsymbol\nu}_{j,k+1}
=
\mathbf J_r(\boldsymbol\xi_{jk}^{+})^{-1}
\boldsymbol\nu_{j,k+1},
$
where $\mathbf J_r$ is the right Jacobian of $\mathrm{SE}(3)$.

WNOA models the local acceleration as zero-mean white noise. Its mean evolution follows constant velocity, while its covariance describes how deviations from that motion accumulate over the interval. The resulting neighboring-state factor is:
\begin{equation}
\begin{aligned}
 \mathbf e_{jk}
 &=
 \begin{bmatrix}
 \boldsymbol\xi_{jk}^{+}
 -\Delta_k\boldsymbol\nu_{jk}
 \\
 \widetilde{\boldsymbol\nu}_{j,k+1}
 -\boldsymbol\nu_{jk}
 \end{bmatrix}, \quad \quad
 \mathcal L_{\mathrm{WNOA}}
 &=
 \frac{1}{3M(S-1)}
 \sum_{j=1}^{M}
 \sum_{k=0}^{S-2}
 \mathbf e_{jk}^{\top}
 \mathbf Q_k^{-1}
 \mathbf e_{jk}.
\end{aligned}
\label{eq:wnoa_main}
\end{equation}

Here, $M$ denotes the number of latent motion nodes. The two blocks of $\mathbf{e}_{jk}$ measure the endpoint pose discrepancy relative to the initial-velocity prediction and the velocity change in common local coordinates, respectively. The analytic covariance $\mathbf Q_k$ contains position, velocity, and cross-covariance blocks that scale respectively with $\Delta_k^3/3$, $\Delta_k$, and $\Delta_k^2/2$.
This time-dependent scaling permits larger motion variation across longer intervals while maintaining tighter coupling between nearby supports. We use separate spectral densities for translational and rotational acceleration and normalize translation using the spatial scale of the
motion scaffold. The complete state-space derivation and the explicit form of $\mathbf Q_k$ are given in Appendix Sec.~\ref{sec:supp_wnoa}.

\paragraph{Continuous state evaluation.}
Track measurements and Gaussian rendering generally occur between support times, so the model must provide a pose at every queried timestamp. For $t\in[\tau_k,\tau_{k+1}]$ and $s=(t-\tau_k)/\Delta_k$, the conditional mean of the WNOA process reduces to cubic Hermite interpolation in the local tangent space:
\begin{equation}
\begin{aligned}
 \overline{\boldsymbol\xi}_{jk}(t)
 &=
 h_{10}(s)\Delta_k\boldsymbol\nu_{jk}
 +h_{01}(s)\boldsymbol\xi_{jk}^{+}
 +h_{11}(s)\Delta_k
 \widetilde{\boldsymbol\nu}_{j,k+1}, \quad \quad
 \mathbf T_j(t) =
 \mathbf T_{jk}
 \operatorname{Exp}\!\left(
 \overline{\boldsymbol\xi}_{jk}(t)^\wedge
 \right),
\end{aligned}
\label{eq:wnoa_query}
\end{equation}
where $h_{10}$, $h_{01}$, and $h_{11}$ are the standard cubic Hermite basis functions detailed in
Appendix Sec.~\ref{sec:supp_wnoa_interpolation}.
Eq.~\eqref{eq:wnoa_query} matches the endpoint poses and body velocities in the local chart.
Measurements from different tracks jointly refine the continuous node trajectories used for Gaussian motion.

\subsection{Hierarchical Gaussian Reconstruction}
\label{sec:hierarchical_motion}

RayCo4D propagates continuous node motion to Gaussian primitives and introduces reference-anchored local corrections for each primitive.
Shared motion captures coherent structure-level dynamics, while local corrections model fine-scale non-rigid deformation. Both components are jointly optimized with Gaussian geometry and appearance under visual and geometric supervision.

\paragraph{Shared motion propagation.}
Each Gaussian $g$ is stored at a reference time $r_g$ with reference pose $\mathbf P_g^{\mathrm{ref}} = [\mathbf R_g^{\mathrm{ref}},\boldsymbol\mu_g^{\mathrm{ref}}]$.
We associate its reference center $\boldsymbol\mu_g^{\mathrm{ref}}$ with a local node set $\mathcal N_g$ using the geometric weighting rule in Eq.~\eqref{eq:ray_association}. These geometric weights provide the initial Gaussian-to-node binding. During photometric fitting, Gaussian-specific logit offsets $b_{gj}$ refine these weights:
\begin{equation}
\begin{aligned}
    \overline a_{gj}
    &=
    \operatorname{softmax}_{j\in\mathcal N_g}
    \left(
        \log a_{gj}^{\mathrm{geo}} + b_{gj}
    \right), \quad \quad
    \mathbf A_g(t)
    =
    \operatorname{DQB}\!\left(
        \left\{
        \overline a_{gj},
        \mathbf T_j(t)\mathbf T_j(r_g)^{-1}
        \right\}_{j\in\mathcal N_g}
    \right),
\end{aligned}
\label{eq:gaussian_binding}
\end{equation}
where $\mathbf A_g(t)\in\mathrm{SE}(3)$ is the shared transformation induced by the continuous node trajectories. The offsets are initialized to zero and activated after the early shared-motion fitting stage. Candidate nodes are refreshed as the scene structure evolves, while the binding for each forward evaluation is computed from the Gaussian's reference geometry. Nearby Gaussians therefore retain overlapping node support and inherit a spatially coherent motion field.

\paragraph{Anchored residual motion (ARM).} 
We equip each dynamic Gaussian with a lightweight local motion curve $\mathbf h_g:[0,1]\rightarrow\mathbb R^6$, represented by a small set of cubic Hermite controls. 
For $\bar t=t/\max(F-1,1)$ and $\bar r_g=r_g/\max(F-1,1)$, anchored residual motion (ARM) is defined as:
\begin{equation}
\begin{aligned}
    \boldsymbol\delta_g(t)
    &=
    \mathbf h_g(\bar t)
    -
    \mathbf h_g(\bar r_g), \qquad
    \mathbf E_g(t)
    =
    \operatorname{Exp}\!\left(
        \left[
        \mathbf D_\ell
        \boldsymbol\delta_g(t)
        \right]^\wedge
    \right),
    \qquad
    \mathbf D_\ell
    =
    \operatorname{diag}
    (\ell\mathbf I_3,\mathbf I_3),
\end{aligned}
\label{eq:anchored_residual}
\end{equation}
where $\ell$ is the spatial unit of the motion scaffold. The translational components of the curve are expressed in scaffold units, and the rotational components are expressed in radians. By construction, $\boldsymbol\delta_g(r_g)=\mathbf 0$ and $\mathbf E_g(r_g)=\mathbf I$ for any control values. The residual is expressed in the Gaussian's own reference coordinate frame and applied before shared motion transport:
\begin{equation}
    \mathbf P_g(t)
    =
    \mathbf A_g(t)\,
    \mathbf P_g^{\mathrm{ref}}\,
    \mathbf E_g(t).
\label{eq:hierarchical_composition}
\end{equation}
In Eq.~\eqref{eq:hierarchical_composition}, $\mathbf E_g(t)$ first adjusts the primitive in its local reference frame, $\mathbf P_g^{\mathrm{ref}}$ places it in the reference scene, and $\mathbf A_g(t)$ transports it through the shared continuous motion. The two identity conditions at $r_g$ preserve the stored reference pose: $\mathbf P_g(r_g)=\mathbf P_g^{\mathrm{ref}}$.
Ray-conditioned track factors supervise the shared node motion encoded by $\mathbf A_g(t)$, while image and surface losses act on the composed Gaussian pose $\mathbf P_g(t)$.
Once the scaffold has stabilized, we enable the local controls and regularize their magnitude, curvature, and neighborhood consistency. The full Hermite parameterization and regularization terms are provided in Appendix Sec.~\ref{sec:supp_local_motion}.

\paragraph{Dynamic rendering and optimization.}
For $\mathbf P_g(t)=[\mathbf R_g(t),\boldsymbol\mu_g(t)]$, the translation and rotation determine the Gaussian center and
orientation at time $t$.
Each primitive additionally maintains its learned scale, opacity, and
spherical-harmonic appearance parameters.
Dynamic Gaussians and a time-independent static Gaussian component are
rendered jointly with differentiable Gaussian
splatting~\citep{kerbl2023gaussians}.
Since both $\mathbf T_j(t)$ and $\mathbf E_g(t)$ admit continuous-time
evaluation, the reconstructed scene can be rendered at any real-valued
time inside the observed interval.
We optimize the scene using the grouped objective:
\begin{equation}
    \mathcal L =
    \mathcal L_{\mathrm{scene}}
    +\lambda_u\mathcal L_{\mathrm{2D}}
    +\lambda_d\mathcal L_{\mathrm{depth}}
    +\lambda_W\mathcal L_{\mathrm{WNOA}}
    +\lambda_A\mathcal L_{\mathrm{ARAP}}
    +\mathcal L_{\mathrm{local}}
    +\mathcal L_{\mathrm{cal}}
    +\lambda_{\mathrm{warp}}
     \mathcal L_{\mathrm{warp}} .
\label{eq:total_objective}
\end{equation}
Here, $\mathcal L_{\mathrm{scene}}$ collects the enabled image reconstruction, rendered-depth, surface-correspondence, and coverage/ownership terms applied to the final rendered scene.
The measurement losses $\mathcal L_{\mathrm{2D}}$ and $\mathcal L_{\mathrm{depth}}$ connect the original track observations to the shared node states, and $\mathcal L_{\mathrm{WNOA}}$ couples those states across support times. The ARAP term preserves the spatial structure of the motion
scaffold~\citep{lei2025mosca}, while $\mathcal L_{\mathrm{local}}$ regularizes the per-Gaussian residual curves. When enabled, $\mathcal{L}_{\text{cal}}$ regularizes camera-pose and depth corrections relative to the input estimates, while $\mathcal{L}_{\text{warp}}$ evaluates adjacent-frame appearance using predicted Gaussian motion. Appendix~\ref{sec:supp_training_objective} provides the expanded definitions of the local and calibration terms.
RayCo4D progressively optimizes the shared motion before introducing the more flexible Gaussian-level variables. Lifted tracks initialize the motion scaffold, with short gaps completed only when constructing candidate node paths. 

\section{Experiments}
\label{sec:experiments}

\subsection{Experimental Setup}
\paragraph{Implementation details.} \label{sec:implementation_details}
We use DA3~\citep{lin2025depth} for camera and depth initialization, BootsTAPIR~\citep{doersch2024bootstap} for long-range tracks, and RAFT~\citep{teed2020raft} for epipolar and adjacent-frame flow-consistency cues.
We optimize with Adam~\citep{kingma2015adam} on a single RTX PRO 6000 Blackwell.
Motion-scaffold and reference-query depth optimization takes 4{,}000 iterations, followed by 8{,}000 iterations of Gaussian warm-up, joint scene--motion refinement, and structure-frozen fitting.
The motion scaffold remains fixed during Gaussian warm-up. Gaussian binding corrections, local residuals, and optional camera--depth calibration are activated by stage.
Pose--velocity supports are placed every two frames, including both sequence endpoints.
The training stages and parameter activation schedule are described in Appendix~\ref{sec:addition_implementation_details}.

\paragraph{Datasets and metrics.}
We evaluate novel-view synthesis on DyCheck~\citep{gao2022dynamic} and NVIDIA~\citep{yoon2020novel}.
DyCheck provides spatially separated iPhone views with limited overlap; NVIDIA provides held-out synchronized camera views from monocular training sequences.
We report peak signal-to-noise ratio (PSNR), structural similarity (SSIM)~\citep{wang2004image}, and learned perceptual image patch similarity (LPIPS)~\citep{zhang2018unreasonable}. On DyCheck, we report mPSNR, mSSIM, and mLPIPS over co-visible regions; NVIDIA follows its standard held-out-view protocol.
We also evaluate motion correspondence on DyCheck using PCK-T, comparing projected 3D motion queries with target-frame point observations.
On NVIDIA, we assess reconstruction at intermediate timestamps withheld during training.
DAVIS~\citep{perazzi2016davis} supplies qualitative tests of rapid motion, occlusion, and non-rigid deformation, as synchronized novel-view ground truth is unavailable.
Dataset and evaluation details are given in Appendix~\ref{sec:dataset_details}.

\paragraph{Baselines.}
We compare with six methods: the shared-motion approaches MoSca~\citep{lei2025mosca} and Shape of Motion~\citep{wang2025shapeofmotion}, the continuous-time baseline SE(3) B-splineGS~\citep{zhang2026se3bspline}, and OriGS~\citep{wu2025origs}, RiGS~\citep{wu2026rigs}, and USPLAT4D~\citep{guo2026usplat4d}.
Additional baseline details are provided in Appendix~\ref{sec:baseline_details}.

\subsection{Comparison with the State of the Art}
\label{sec:comparison}

\begin{table}[t]
\centering
\caption{
\textbf{Novel-view synthesis and component ablations.}
DyCheck metrics are computed within the co-visible regions, while NVIDIA
metrics follow the standard held-out-view protocol. The lower block reports component ablations on both datasets; underlines mark the best baseline.
}
\label{tab:main_results}
\small
\setlength{\tabcolsep}{4.5pt}
\renewcommand{\arraystretch}{1.12}
\begin{tabular}{@{}p{0.30\linewidth}*{6}{>{\centering\arraybackslash}p{\dimexpr0.70\linewidth/6-2\tabcolsep\relax}}@{}}
\toprule
& \multicolumn{3}{c}{\textbf{DyCheck}}
& \multicolumn{3}{c}{\textbf{NVIDIA}} \\
\cmidrule(lr){2-4}
\cmidrule(lr){5-7}
\textbf{Method}
& mPSNR$\uparrow$
& mSSIM$\uparrow$
& mLPIPS$\downarrow$
& PSNR$\uparrow$
& SSIM$\uparrow$
& LPIPS$\downarrow$ \\
\midrule
MoSca~\citep{lei2025mosca}
& 19.86 & 0.702 & 0.298
& 25.89 & 0.839 & 0.079 \\

Shape of Motion~\citep{wang2025shapeofmotion}
& 18.63 & 0.674 & 0.321
& 24.89 & 0.732 & 0.113 \\

OriGS~\citep{wu2025origs}
& 19.97 & 0.707 & 0.307
& 26.53 & 0.850 & 0.068 \\

RiGS~\citep{wu2026rigs}
& 18.89 & 0.689 & 0.354
& 25.24 & 0.793 & 0.098 \\

SE(3) B-splineGS~\citep{zhang2026se3bspline}
& 20.53 & 0.713 & 0.276
& 26.78 & 0.848 & 0.063 \\

USPLAT4D~\citep{guo2026usplat4d}
& \underline{20.64}
& \underline{0.721}
& \underline{0.263}
& \underline{26.99}
& \underline{0.853}
& \underline{0.059} \\

\midrule
RayCo4D (w/o MSD) & 19.62 & 0.691 & 0.313 & 26.61 & 0.845 & 0.062 \\
RayCo4D (w/o RCQ) & 19.78 & 0.699 & 0.299 & 26.72 & 0.851 & 0.057 \\
RayCo4D (w/o WNOA factor) & 19.94 & 0.703 & 0.283 & 26.77 & 0.854 & 0.055 \\
RayCo4D (w/o ARM) & 20.79 & 0.738 & 0.247 & 27.77 & 0.868 & 0.049 \\
\rowcolor{black!4}
\textbf{RayCo4D} (Full)
& \textbf{20.97}
& \textbf{0.743}
& \textbf{0.241}
& \textbf{27.94}
& \textbf{0.872}
& \textbf{0.045} \\ \bottomrule \end{tabular}
\end{table}

\paragraph{Quantitative comparison.}
Tab.~\ref{tab:main_results} compares RayCo4D with six representative
baselines on DyCheck and NVIDIA.
RayCo4D achieves the best results across all reported metrics
on both datasets.
Compared with USPLAT4D, the strongest baseline,
RayCo4D improves DyCheck mPSNR by 0.33 dB and mSSIM by 0.022,
while reducing mLPIPS by 0.022 (8.4\%).
On NVIDIA, it achieves larger gains in PSNR (+0.95 dB)
and perceptual quality, reducing LPIPS by 0.014 (23.7\%),
alongside a 0.019 improvement in SSIM.
These consistent improvements across fidelity, structural similarity,
and perceptual metrics demonstrate the effectiveness of RayCo4D
under both evaluation protocols.

\begin{figure*}[t]
\centering
\includegraphics[width=\textwidth]{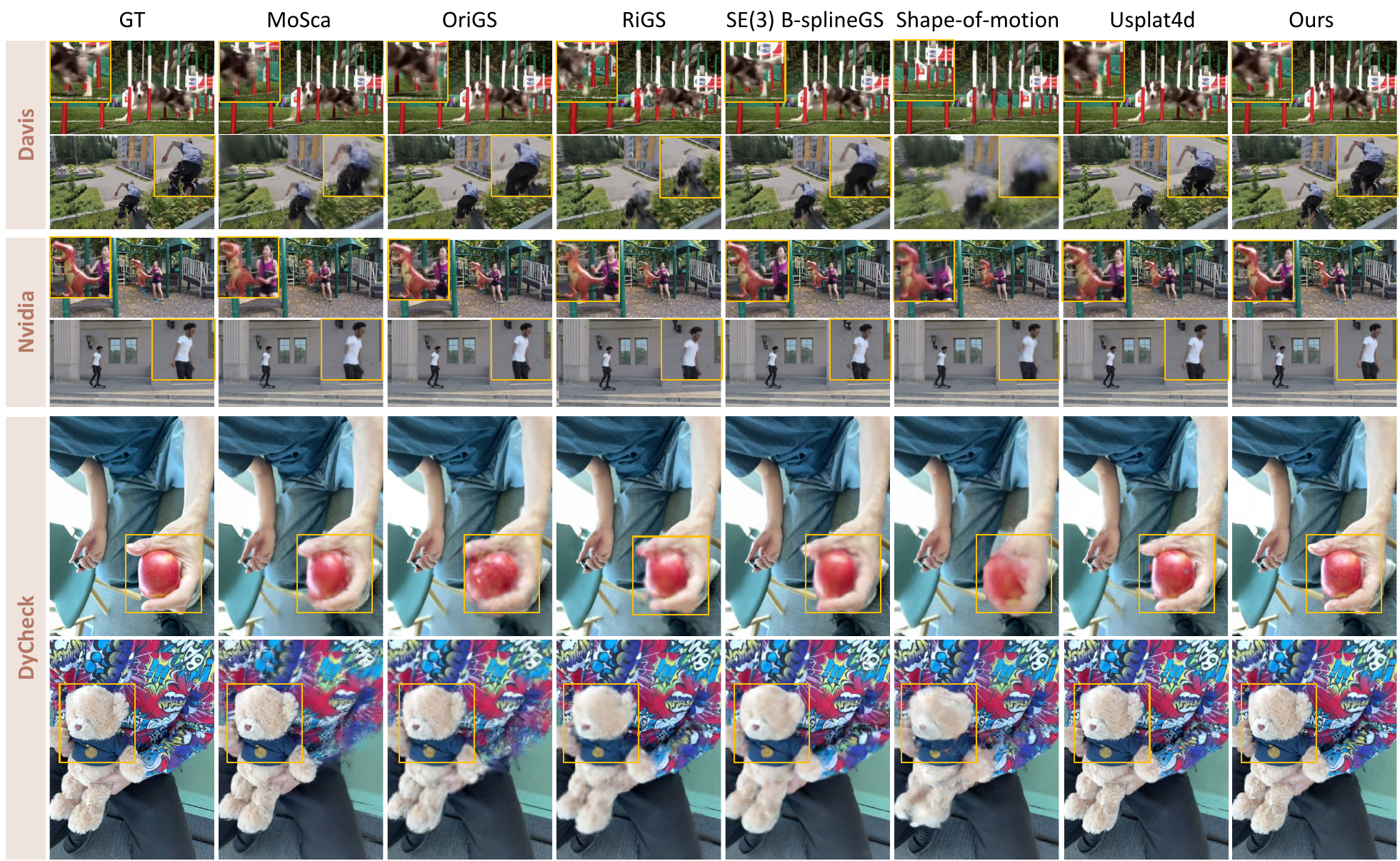}
    \vspace{-0.3cm}
\caption{
\textbf{Qualitative comparisons on DyCheck, NVIDIA, and DAVIS.}
Ground-truth held-out views are shown for DyCheck and NVIDIA; DAVIS provides qualitative comparisons under rapid motion, occlusion, and non-rigid deformation.
}
\label{fig:qualitative}
\end{figure*}

\paragraph{Qualitative comparison.}
Fig.~\ref{fig:qualitative} presents qualitative comparisons
on DyCheck, NVIDIA, and DAVIS. On DyCheck and NVIDIA, RayCo4D produces sharper object boundaries and more faithful fine details, with reduced blurring and ghosting in dynamic regions.
The magnified regions highlight improvements around moving human silhouettes, object contours, and fine structures, where competing methods often exhibit motion artifacts or loss of detail. On DAVIS, RayCo4D further maintains coherent object
shapes and clearer boundaries under rapid motion, partial occlusion, and non-rigid deformation. These results complement the quantitative evaluation and illustrate the method's performance on challenging real-world dynamic scenes.

\paragraph{Efficiency comparison.}
In terms of efficiency, RayCo4D requires 86 minutes on 200-frame DyCheck sequences, including 36 minutes for preprocessing and 50 minutes for training. Its overall runtime is comparable to the evaluated baselines. Appendix~\ref{sec:supp_runtime} details the timing protocol and provides the complete per-method preprocessing and training breakdown in Tab.~\ref{tab:supp_runtime}.

\paragraph{Failure cases.}
RayCo4D may struggle when motion changes abruptly during long observation gaps or when geometric estimates are unreliable. These challenges can lead to motion errors, geometric inconsistencies, and visible reconstruction artifacts, as shown in Appendix Sec.~\ref{sec:supp_challenging_cases}.

\subsection{Dissecting RayCo4D: Measurements, Dynamics, and Local Motion}
\label{sec:analysis}
We evaluate component contributions through quantitative ablations on DyCheck and NVIDIA.
All ablations use the same input observations, initialization, training budget, and evaluation protocol as the full model, with only the indicated component modified.

\begin{figure}[H]
\centering
\includegraphics[width=\linewidth]{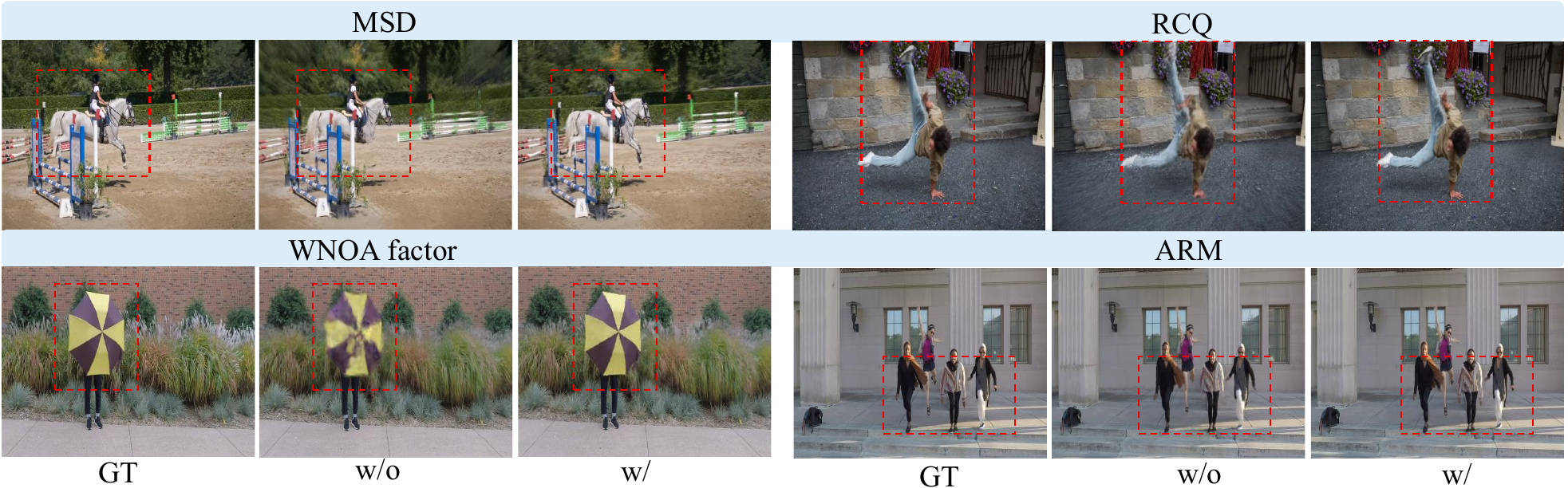}
\vspace{-0.3cm}
\caption{\textbf{Qualitative ablations on NVIDIA and DAVIS.}
Comparisons with the full model illustrate the effects of individual components on structural coherence and fine-scale deformation.}
\label{fig:ablation_qualitative}
\end{figure}

\paragraph{What drives reconstruction quality?}
Tab.~\ref{tab:main_results} evaluates four design choices.
\textbf{w/o MSD} replaces the many-to-many measurement--state
association with track-specific motion support;
\textbf{w/o RCQ} optimizes reference queries freely in $\mathbb R^3$
instead of along their camera rays;
\textbf{w/o WNOA factor} removes the process coupling between
neighboring pose--velocity states; and
\textbf{w/o ARM} disables the reference-anchored local residual.
Removing MSD causes the largest mPSNR/PSNR drops:
$1.35$\,dB on DyCheck and $1.33$\,dB on NVIDIA,
with a $37.8\%$ increase in NVIDIA LPIPS relative to the full model.
The less distinct horse limb boundaries in
Fig.~\ref{fig:ablation_qualitative} further illustrate the
importance of shared motion support.
Removing RCQ or the WNOA factor lowers mPSNR/PSNR by more
than $1$\,dB on both datasets and increases NVIDIA LPIPS
by $26.7\%$ and $22.2\%$, respectively.
Their visual effects are complementary: removing RCQ blurs
the dancer's extended limbs, whereas removing the WNOA factor
blurs the umbrella's outline and colored panel boundaries.
Together, these results support the roles of ray constraints
and temporal coupling in recovering coherent shared geometry.
Removing ARM has a smaller effect on whole-image fidelity,
reducing mPSNR/PSNR by $0.18$/$0.17$\,dB on DyCheck/NVIDIA,
but still increases mLPIPS/LPIPS by $2.5\%$/$8.9\%$.
The sharper limb contours in the jumping example with ARM
illustrate its contribution to localized deformation
beyond the shared motion scaffold.

\begingroup
\setlength{\intextsep}{0pt}
\Needspace{12\baselineskip}
\begin{wraptable}{r}{0.50\textwidth}
\centering
\setlength{\abovecaptionskip}{0pt}
\caption{\textbf{Motion correspondence on DyCheck (PCK-T$\uparrow$).} Best result is highlighted in bold.}
\label{tab:rayco4d_correspondence}
\normalsize
\setlength{\tabcolsep}{1pt}
\renewcommand{\arraystretch}{1.05}
\begin{tabular*}{\linewidth}{@{\extracolsep{\fill}}cccc@{}}
\toprule
{\renewcommand{\arraystretch}{0.75}\begin{tabular}[c]{@{}c@{}}MoSca\\{\footnotesize (\citeauthor{lei2025mosca},}\\{\footnotesize \citeyear{lei2025mosca})}\end{tabular}}
& {\renewcommand{\arraystretch}{0.75}\begin{tabular}[c]{@{}c@{}}OriGS\\{\footnotesize (\citeauthor{wu2025origs},}\\{\footnotesize \citeyear{wu2025origs})}\end{tabular}}
& {\renewcommand{\arraystretch}{0.75}\begin{tabular}[c]{@{}c@{}}SE(3) B-splineGS\\{\footnotesize (\citeauthor{zhang2026se3bspline},}\\{\footnotesize \citeyear{zhang2026se3bspline})}\end{tabular}}
& \textbf{RayCo4D} \\
\midrule
0.822 & 0.826 & \underline{0.832} & \textbf{0.840} \\
\bottomrule
\end{tabular*}
\end{wraptable}
\paragraph{Motion correspondence.}
Tab.~\ref{tab:rayco4d_correspondence} evaluates motion correspondence on DyCheck by projecting reconstructed 3D motion queries into target frames. RayCo4D achieves a PCK-T of $0.840$, yielding relative improvements of $1.0\%$ over SE(3) B-splineGS and $2.2\%$ over MoSca. These gains suggest that the shared motion representation more accurately transports scene points across frames, complementing the rendering results with evidence of improved motion correspondence. This finding is consistent with the role of shared latent states in integrating track observations across different visibility intervals.

\par
\endgroup

\begingroup
\setlength{\intextsep}{0pt}
\Needspace{12\baselineskip}
\begin{wraptable}{r}{0.50\textwidth}
\centering
\setlength{\abovecaptionskip}{0pt}
\caption{\textbf{Intermediate-time reconstruction on NVIDIA.}
Best result is highlighted in bold.}
\label{tab:rayco4d_intermediate}
\normalsize
\setlength{\tabcolsep}{2pt}
\renewcommand{\arraystretch}{1.05}
\begin{tabular*}{\linewidth}{@{\extracolsep{\fill}}lccc@{}}
\toprule
\textbf{Method} & PSNR$\uparrow$ & SSIM$\uparrow$ & LPIPS$\downarrow$ \\
\midrule
\begin{tabular}[t]{@{}l@{}}SE(3) B-splineGS\\\citep{zhang2026se3bspline}\end{tabular} & \underline{25.76} & \underline{0.835} & \underline{0.081} \\
\textbf{RayCo4D} & \textbf{26.87} & \textbf{0.851} & \textbf{0.060} \\
\bottomrule
\end{tabular*}
\end{wraptable}
\paragraph{Intermediate-time reconstruction.}
Tab.~\ref{tab:rayco4d_intermediate} evaluates NVIDIA intermediate frames withheld from image, track, and depth supervision during training. Under the same temporal subsampling protocol, RayCo4D outperforms SE(3) B-splineGS by $1.11$\,dB in PSNR, increases SSIM by $1.9\%$, and reduces LPIPS by $25.9\%$. These gains indicate improved reconstruction at unobserved timestamps, particularly in perceptual quality.
\par
\endgroup

\section{Conclusion}
\label{sec:conclusion}
We presented \textbf{RayCo4D}, a ray-conditioned continuous motion model for monocular 4D reconstruction. RayCo4D utilizes 2D tracks together with camera-depth geometry to constrain shared latent motion nodes through local many-to-many associations. Sparse $\mathrm{SE}(3)$ pose--velocity states, coupled by a white-noise-on-acceleration process prior, form continuous node trajectories that integrate evidence across different visibility intervals. The inferred motion is propagated to dynamic Gaussians through dual-quaternion blending, while reference-anchored local $\mathrm{SE}(3)$ corrections capture fine-grained non-rigid deformation. Experiments on DyCheck and NVIDIA scenes demonstrate competitive novel-view synthesis quality and improved novel-view synthesis, motion correspondence, and intermediate-time reconstruction. 
Our results validate that separating visual measurements from latent motion states provides a coherent and robust foundation for monocular 4D reconstruction.

\bibliography{iclr2027_conference}
\bibliographystyle{iclr2027_conference}

\newpage
\appendix


\section*{Supplementary Material}
This supplementary material provides additional details and results for RayCo4D.
Sections A and B present additional related work and notation.
Section C derives the WNOA formulation and its continuous-time interpolation.
Section D describes hierarchical Gaussian reconstruction, training, evaluation protocols, and runtime.
Sections E and F present additional experimental results, challenging cases, and scene-wise qualitative comparisons.

\section{Additional Related Work}
\label{sec:additional_related_work}

\paragraph{Spatial structure and motion sharing.}
Recent structured representations address different aspects of dynamic
scene complexity.
OriGS~\citep{wu2025origs} uses a Global Orientation Field to organize
region-specific deformations.
RiGS~\citep{wu2026rigs} separates static, rigid, and transient Gaussians
to accommodate motion at different temporal scales.
USPLAT4D~\citep{guo2026usplat4d} constructs a spatio-temporal motion
graph using per-Gaussian uncertainty, propagating reliable motion cues
toward weakly observed regions.
RayCo4D organizes spatial sharing through local track--node
associations: reference queries retain their geometric parameters,
while shared node states encode time-varying motion.

\paragraph{Spline trajectories and local refinement.}
Beyond SplineGS~\citep{park2025splinegs} and
SE(3) B-splineGS~\citep{zhang2026se3bspline},
WebSpline~\citep{park2026webspline} initializes a Structural Proxy
Graph from 2D tracks to regularize per-Gaussian Hermite trajectories.
RetimeGS~\citep{wang2026retimegs} defines continuous-time Gaussian
behavior for temporal retiming and intermediate-time rendering.
RayCo4D also uses cubic Hermite expressions, at two distinct levels:
the WNOA conditional mean interpolates shared pose--velocity states
in a local $\mathrm{SE}(3)$ chart, and per-Gaussian Hermite curves
parameterize reference-anchored local corrections.
The shared process factors and local residual constraints determine
how these temporal representations are estimated and combined.

\paragraph{Probabilistic and dynamical formulations.}
GP-4DGS~\citep{kim2026gp4dgs} models spatio-temporal deformation
correlations through variational Gaussian processes, supporting
uncertainty estimation and temporal extrapolation.
Gaussians on their Way~\citep{deng2025gaussians} treats Gaussian means
and covariances as dynamic states and fuses motion predictions with
deformation observations.
EvoGS~\citep{asiimwe2025evogs} integrates a learned velocity field,
while ODE-GS~\citep{wang2026odegs} evolves latent states derived from
reconstructed Gaussian trajectories through a Neural ODE.
RayCo4D places its WNOA prior on sparse shared
$\mathrm{SE}(3)$ pose--velocity states.
Ray-conditioned measurements constrain these states through local
associations, and their continuous transformations drive the dense
Gaussian representation.

\section{Notation}
\label{sec:supp_notation}

Tab.~\ref{tab:supp_notation} summarizes the main symbols.
Time $t\in[0,F-1]$ is measured in frame units, with noninteger
values denoting subframe times.
All rigid transformations act on column vectors:
$\mathbf T\cdot\mathbf x=\mathbf R\mathbf x+\mathbf p$
for $\mathbf T=[\mathbf R,\mathbf p]\in\mathrm{SE}(3)$,
and transformation products act from right to left.
Twists are ordered as translation followed by rotation,
with body velocity defined by
$\dot{\mathbf T}(t)=\mathbf T(t)\boldsymbol\nu(t)^\wedge$.
The operators $(\cdot)^\wedge$ and $(\cdot)^\vee$ convert
between six-dimensional vectors and $\mathfrak{se}(3)$ matrices;
$\operatorname{Exp}$ and $\operatorname{Log}$ denote the
group exponential and local logarithm.

For camera-to-world pose
$\mathbf C_f=[\mathbf R_f^c,\mathbf c_f]$
and intrinsics $\mathbf K_f$, the image projection and
camera-axis depth of a world point $\mathbf x$ are
\begin{equation}
\begin{aligned}
\pi_f(\mathbf x)
&=\pi\!\left(
\mathbf K_f(\mathbf R_f^c)^\top(\mathbf x-\mathbf c_f)
\right),\\
z_f(\mathbf x)
&=\mathbf e_3^\top(\mathbf R_f^c)^\top
(\mathbf x-\mathbf c_f),
\end{aligned}
\end{equation}
where $\pi([x,y,z]^\top)=[x/z,y/z]^\top$ and
$\mathbf e_3=[0,0,1]^\top$.
Depth denotes the camera-axis coordinate rather than
Euclidean distance to the camera center.

\begin{table}[t]
\centering
\caption{\textbf{Main notation used in RayCo4D.}
Symbols are grouped by their roles in the representation.}
\label{tab:supp_notation}
\small
\setlength{\tabcolsep}{5pt}
\renewcommand{\arraystretch}{1.08}
\begin{tabular}{
@{}p{0.29\linewidth}
p{\dimexpr0.71\linewidth-2\tabcolsep\relax}@{}
}
\toprule
\textbf{Symbol} & \textbf{Meaning} \\
\midrule
\multicolumn{2}{@{}l}{\textit{Indices and time}} \\
$f,i,j,g,k$
& Frame, track, motion-node, Gaussian, and support-time indices. \\
$F,M,S$
& Numbers of video frames, motion nodes, and temporal supports. \\
$t,\tau_k,\Delta_k$
& Query time, support time, and interval
$\Delta_k=\tau_{k+1}-\tau_k$. \\
$r_i,r_g$
& Reference times of track $i$ and Gaussian $g$. \\
$\bar t$
& Normalized time $t/\max(F-1,1)$ for local motion curves. \\

\midrule
\multicolumn{2}{@{}l}{\textit{Measurements and associations}} \\
$\mathbf C_f,\mathbf K_f$
& Camera-to-world pose and camera intrinsics. \\
$\mathbf u_{if},c_{if}$
& Observed track pixel and its confidence. \\
$\widetilde{\mathbf u}_{if}$
& Homogeneous pixel coordinate
$[\mathbf u_{if}^{\top},1]^{\top}$. \\
$\mathcal O_u,\mathcal O_d$
& Visible observations and the subset with valid depth. \\
$\mathbf q_i,\widehat{\mathbf x}_i(t)$
& Reference query in world space and its predicted position at $t$. \\
$d_i^0,z_i,\eta_i$
& Initial query depth, optimized camera-axis depth,
and scalar depth parameter. \\
$\mathcal N_i,a_{ij}$
& Track-to-node neighborhood and normalized association weight. \\
$K,\sigma_j$
& Number of neighboring nodes and node influence radius. \\

\midrule
\multicolumn{2}{@{}l}{\textit{Continuous shared motion}} \\
$\mathbf T_j(t)$
& Continuous rigid pose of motion node $j$. \\
$\mathbf T_{jk},\boldsymbol\nu_{jk}$
& Node pose and body-frame velocity at $\tau_k$. \\
$\boldsymbol\xi_{jk}(t)$
& Local pose coordinate
$\operatorname{Log}(\mathbf T_{jk}^{-1}\mathbf T_j(t))^\vee$. \\
$\mathbf J_r$
& Right Jacobian of $\mathrm{SE}(3)$. \\
$\mathbf e_{jk}$
& Neighboring-state WNOA residual in $\mathbb R^{12}$. \\
$\mathbf Q_c,\mathbf Q_k$
& Acceleration spectral-density matrix and integrated
process covariance over $\Delta_k$. \\
$\ell$
& Characteristic spatial scale of the motion scaffold. \\

\midrule
\multicolumn{2}{@{}l}{\textit{Hierarchical Gaussian motion}} \\
$\mathbf P_g^{\mathrm{ref}},\mathbf P_g(t)$
& Reference Gaussian pose and composed pose at time $t$. \\
$\mathcal N_g,\overline a_{gj}$
& Gaussian-to-node neighborhood and normalized binding weight. \\
$a_{gj}^{\mathrm{geo}},b_{gj}$
& Geometric binding weight and learnable logit offset. \\
$\operatorname{DQB}$
& Dual-quaternion blending of rigid transformations. \\
$\mathbf A_g(t),\mathbf E_g(t)$
& Shared motion transform and reference-anchored local residual. \\
$\mathbf h_g,\boldsymbol\delta_g(t)$
& Local motion curve and anchored increment
$\mathbf h_g(\bar t)-\mathbf h_g(\bar r_g)$. \\
$\mathbf D_\ell$
& Translation-scaling matrix
$\operatorname{diag}(\ell\mathbf I_3,\mathbf I_3)$. \\
\bottomrule
\end{tabular}
\end{table}

\section{Full WNOA Formulation}
\label{sec:supp_wnoa}

This section provides the complete formulation of the continuous-time
white-noise-on-acceleration process used in RayCo4D.
Twists are ordered as translation followed by rotation,
body velocities satisfy
$\dot{\mathbf T}(t)=\mathbf T(t)\boldsymbol\nu(t)^\wedge$,
and time is measured in video-frame units.

\subsection{Local Coordinates on \texorpdfstring{$\mathrm{SE}(3)$}{SE(3)}}

Consider node $j$ over the support interval
$[\tau_k,\tau_{k+1}]$.
We define a local tangent-space trajectory centered at the left endpoint:
\begin{equation}
 \boldsymbol\xi_{jk}(t)
 =
 \operatorname{Log}\!\left(
 \mathbf T_{jk}^{-1}\mathbf T_j(t)
 \right)^\vee.
\label{eq:supp_local_coordinate}
\end{equation}
The derivative of this local coordinate is related to the body-frame
velocity through the right Jacobian:
\begin{equation}
 \dot{\boldsymbol\xi}_{jk}(t)
 =
 \mathbf J_r\!\left(
 \boldsymbol\xi_{jk}(t)
 \right)^{-1}
 \boldsymbol\nu_j(t).
\label{eq:supp_local_velocity}
\end{equation}

At the left endpoint,
$\boldsymbol\xi_{jk}(\tau_k)=\mathbf 0$, and hence
\begin{equation}
 \dot{\boldsymbol\xi}_{jk}(\tau_k)
 =
 \boldsymbol\nu_{jk}.
\end{equation}
At the right endpoint, define
\begin{equation}
\begin{aligned}
 \boldsymbol\xi_{jk}^{+}
 &=
 \operatorname{Log}\!\left(
 \mathbf T_{jk}^{-1}\mathbf T_{j,k+1}
 \right)^\vee,
 \\
 \widetilde{\boldsymbol\nu}_{j,k+1}
 &=
 \mathbf J_r(\boldsymbol\xi_{jk}^{+})^{-1}
 \boldsymbol\nu_{j,k+1}.
\end{aligned}
\label{eq:supp_endpoint_state}
\end{equation}
The transformed velocity
$\widetilde{\boldsymbol\nu}_{j,k+1}$
is the derivative of the local coordinate at the right endpoint.
It allows both endpoint velocities to be compared in the same tangent
space.

\subsection{Linear WNOA Process in the Local Chart}

We concatenate local pose and velocity into the state
\begin{equation}
 \boldsymbol\gamma_{jk}(t)
 =
 \begin{bmatrix}
 \boldsymbol\xi_{jk}(t)
 \\
 \dot{\boldsymbol\xi}_{jk}(t)
 \end{bmatrix}
 \in\mathbb R^{12}.
\end{equation}
The WNOA process assumes
\begin{equation}
 \ddot{\boldsymbol\xi}_{jk}(t)
 =
 \mathbf w(t),
 \qquad
 \mathbb E[
 \mathbf w(t)\mathbf w(t')^\top]
 =
 \mathbf Q_c\delta(t-t'),
\label{eq:supp_wnoa_noise}
\end{equation}
which gives the linear time-invariant stochastic differential equation
\begin{equation}
 \dot{\boldsymbol\gamma}_{jk}(t)
 =
 \underbrace{
 \begin{bmatrix}
 \mathbf 0 & \mathbf I_6
 \\
 \mathbf 0 & \mathbf 0
 \end{bmatrix}}_{\mathbf A}
 \boldsymbol\gamma_{jk}(t)
 +
 \underbrace{
 \begin{bmatrix}
 \mathbf 0
 \\
 \mathbf I_6
 \end{bmatrix}}_{\mathbf L}
 \mathbf w(t).
\label{eq:supp_wnoa_sde}
\end{equation}

For a time interval of length $\delta$, the state-transition matrix is
\begin{equation}
 \boldsymbol\Phi(\delta)
 =
 \exp(\mathbf A\delta)
 =
 \begin{bmatrix}
 \mathbf I_6 & \delta\mathbf I_6
 \\
 \mathbf 0   & \mathbf I_6
 \end{bmatrix}.
\label{eq:supp_transition}
\end{equation}
Integrating the white acceleration noise gives the process covariance
\begin{equation}
\begin{aligned}
 \mathbf Q(\delta)
 &=
 \int_0^\delta
 \boldsymbol\Phi(\delta-s)
 \mathbf L\mathbf Q_c\mathbf L^\top
 \boldsymbol\Phi(\delta-s)^\top
 \,\mathrm ds
 \\
 &=
 \begin{bmatrix}
 \frac{\delta^3}{3}\mathbf Q_c
 &
 \frac{\delta^2}{2}\mathbf Q_c
 \\
 \frac{\delta^2}{2}\mathbf Q_c
 &
 \delta\mathbf Q_c
 \end{bmatrix}.
\end{aligned}
\label{eq:supp_process_covariance}
\end{equation}

We use separate acceleration spectral densities for translation and
rotation:
\begin{equation}
 \mathbf Q_c
 =
 \operatorname{diag}
 \left(
 q_p\ell^2\mathbf I_3,
 q_R\mathbf I_3
 \right),
\label{eq:supp_spectral_density}
\end{equation}
where $\ell$ is the characteristic spatial scale of the motion scaffold.
Equivalently, translational pose and velocity residuals can be normalized
by $\ell$ before applying the density $q_p$.

The endpoint states in the local chart are
\begin{equation}
 \boldsymbol\gamma_{jk}^{-}
 =
 \begin{bmatrix}
 \mathbf 0
 \\
 \boldsymbol\nu_{jk}
 \end{bmatrix},
 \qquad
 \boldsymbol\gamma_{jk}^{+}
 =
 \begin{bmatrix}
 \boldsymbol\xi_{jk}^{+}
 \\
 \widetilde{\boldsymbol\nu}_{j,k+1}
 \end{bmatrix}.
\end{equation}
For $\Delta_k=\tau_{k+1}-\tau_k$, the transition residual becomes
\begin{equation}
\begin{aligned}
 \mathbf e_{jk}
 &=
 \boldsymbol\gamma_{jk}^{+}
 -
 \boldsymbol\Phi(\Delta_k)
 \boldsymbol\gamma_{jk}^{-}
 \\
 &=
 \begin{bmatrix}
 \boldsymbol\xi_{jk}^{+}
 -\Delta_k\boldsymbol\nu_{jk}
 \\
 \widetilde{\boldsymbol\nu}_{j,k+1}
 -\boldsymbol\nu_{jk}
 \end{bmatrix}.
\end{aligned}
\label{eq:supp_wnoa_residual}
\end{equation}
The corresponding process objective is
\begin{equation}
 \mathcal L_{\mathrm{WNOA}}
 =
 \frac{1}{3M(S-1)}
 \sum_{j=1}^{M}
 \sum_{k=0}^{S-2}
 \mathbf e_{jk}^{\top}
 \mathbf Q(\Delta_k)^{-1}
 \mathbf e_{jk}.
\label{eq:supp_wnoa_loss}
\end{equation}

\paragraph{Equivalent implementation form.}
Let
\begin{equation}
 \mathbf e_{jk}^{p}
 =
 \boldsymbol\xi_{jk}^{+}
 -\Delta_k\boldsymbol\nu_{jk},
 \qquad
 \mathbf e_{jk}^{v}
 =
 \widetilde{\boldsymbol\nu}_{j,k+1}
 -\boldsymbol\nu_{jk}.
\end{equation}
The full covariance-weighted quadratic can be evaluated without explicitly
forming $\mathbf Q(\Delta_k)^{-1}$.
For translation, the equivalent energy is
\begin{equation}
 \mathcal E_{jk}^{\mathrm{trans}}
 =
 \frac{12}{q_p\Delta_k^3}
 \left\|
 \frac{
 \mathbf e_{jk,1:3}^{p}
 -\frac{\Delta_k}{2}\mathbf e_{jk,1:3}^{v}
 }{\ell}
 \right\|_2^2
 +
 \frac{1}{q_p\Delta_k}
 \left\|
 \frac{\mathbf e_{jk,1:3}^{v}}{\ell}
 \right\|_2^2,
\label{eq:supp_wnoa_translation}
\end{equation}
and the rotational energy is
\begin{equation}
 \mathcal E_{jk}^{\mathrm{rot}}
 =
 \frac{12}{q_R\Delta_k^3}
 \left\|
 \mathbf e_{jk,4:6}^{p}
 -\frac{\Delta_k}{2}\mathbf e_{jk,4:6}^{v}
 \right\|_2^2
 +
 \frac{1}{q_R\Delta_k}
 \left\|
 \mathbf e_{jk,4:6}^{v}
 \right\|_2^2.
\label{eq:supp_wnoa_rotation}
\end{equation}
Equations~\eqref{eq:supp_wnoa_translation} and
\eqref{eq:supp_wnoa_rotation} are algebraically equivalent to the quadratic
form in Eq.~\eqref{eq:supp_wnoa_loss}.

\subsection{Continuous-Time Conditional Mean}
\label{sec:supp_wnoa_interpolation}

For a query time
$t\in[\tau_k,\tau_{k+1}]$, let
\begin{equation}
 \delta=t-\tau_k,
 \qquad
 \Delta_k=\tau_{k+1}-\tau_k,
 \qquad
 s=\frac{\delta}{\Delta_k}.
\end{equation}
The conditional mean of the linear WNOA process can be written as
\begin{equation}
\begin{aligned}
 \overline{\boldsymbol\gamma}_{jk}(t)
 &=
 \boldsymbol\Phi(\delta)
 \boldsymbol\gamma_{jk}^{-}
 \\
 &\quad+
 \mathbf Q(\delta)
 \boldsymbol\Phi(\Delta_k-\delta)^\top
 \mathbf Q(\Delta_k)^{-1}
 \left(
 \boldsymbol\gamma_{jk}^{+}
 -
 \boldsymbol\Phi(\Delta_k)
 \boldsymbol\gamma_{jk}^{-}
 \right).
\end{aligned}
\label{eq:supp_gp_interpolation}
\end{equation}

For WNOA, the pose component of
Eq.~\eqref{eq:supp_gp_interpolation}
reduces to cubic Hermite interpolation.
The basis functions are
\begin{equation}
\begin{aligned}
 h_{00}(s)&=2s^3-3s^2+1,
&
 h_{10}(s)&=s^3-2s^2+s,
\\
 h_{01}(s)&=-2s^3+3s^2,
&
 h_{11}(s)&=s^3-s^2.
\end{aligned}
\label{eq:supp_hermite_basis}
\end{equation}
Because the local pose at the left endpoint is zero,
the $h_{00}$ term vanishes, giving
\begin{equation}
 \overline{\boldsymbol\xi}_{jk}(t)
 =
 h_{10}(s)\Delta_k\boldsymbol\nu_{jk}
 +
 h_{01}(s)\boldsymbol\xi_{jk}^{+}
 +
 h_{11}(s)\Delta_k
 \widetilde{\boldsymbol\nu}_{j,k+1}.
\label{eq:supp_hermite_pose}
\end{equation}
The continuous pose is then recovered on the group:
\begin{equation}
 \mathbf T_j(t)
 =
 \mathbf T_{jk}
 \operatorname{Exp}\!\left(
 \overline{\boldsymbol\xi}_{jk}(t)^\wedge
 \right).
\label{eq:supp_group_pose}
\end{equation}

For completeness, differentiating the Hermite curve gives
\begin{equation}
\begin{aligned}
 \dot{\overline{\boldsymbol\xi}}_{jk}(t)
 &=
 \frac{1}{\Delta_k}
 \Bigl[
 h_{10}'(s)\Delta_k\boldsymbol\nu_{jk}
 +
 h_{01}'(s)\boldsymbol\xi_{jk}^{+}
 \\
 &\hspace{4.7em}
 +
 h_{11}'(s)\Delta_k
 \widetilde{\boldsymbol\nu}_{j,k+1}
 \Bigr],
\end{aligned}
\label{eq:supp_hermite_velocity}
\end{equation}
and the corresponding body velocity is
\begin{equation}
 \boldsymbol\nu_j(t)
 =
 \mathbf J_r\!\left(
 \overline{\boldsymbol\xi}_{jk}(t)
 \right)
 \dot{\overline{\boldsymbol\xi}}_{jk}(t).
\label{eq:supp_body_velocity}
\end{equation}
This construction satisfies
\begin{equation}
\begin{aligned}
 \mathbf T_j(\tau_k)&=\mathbf T_{jk},
&
 \mathbf T_j(\tau_{k+1})&=\mathbf T_{j,k+1},
\\
 \boldsymbol\nu_j(\tau_k)&=\boldsymbol\nu_{jk},
&
 \boldsymbol\nu_j(\tau_{k+1})&=\boldsymbol\nu_{j,k+1}.
\end{aligned}
\end{equation}
Thus, the process factor and the continuous pose query are derived from the
same pose--velocity state model.

\section{Additional Details of Hierarchical Gaussian Reconstruction}
\label{sec:supp_hierarchical_reconstruction}

\subsection{Gaussian-to-Node Binding}
\label{sec:supp_gaussian_binding}

For Gaussian $g$, let
$\boldsymbol\mu_g^{\mathrm{ref}}$
be its reference center and $\mathcal N_g$ its $K$ nearest motion nodes
at reference time $r_g$.
The geometric binding weights are
\begin{equation}
    a_{gj}^{\mathrm{geo}}
    =
    \frac{
    \exp\!\left(
    -\|
    \boldsymbol\mu_g^{\mathrm{ref}}
    -
    \mathbf p_j(r_g)
    \|_2^2/(2\sigma_j^2)
    \right)
    }{
    \sum_{\ell\in\mathcal N_g}
    \exp\!\left(
    -\|
    \boldsymbol\mu_g^{\mathrm{ref}}
    -
    \mathbf p_\ell(r_g)
    \|_2^2/(2\sigma_\ell^2)
    \right)
    }.
\label{eq:supp_gaussian_geometric_weights}
\end{equation}
Photometric fitting introduces the Gaussian-specific offsets
$b_{gj}$:
\begin{equation}
    \overline a_{gj}
    =
    \frac{
        a_{gj}^{\mathrm{geo}}\exp(b_{gj})
    }{
        \sum_{\ell\in\mathcal N_g}
        a_{g\ell}^{\mathrm{geo}}\exp(b_{g\ell})
    }.
\label{eq:supp_gaussian_corrected_weights}
\end{equation}
The offsets are initialized to zero, so
$\overline a_{gj}=a_{gj}^{\mathrm{geo}}$
at the beginning of optimization.

For each node, define the relative transformation from the Gaussian
reference time to query time $t$:
\begin{equation}
    \Delta\mathbf T_{gj}(t)
    =
    \mathbf T_j(t)\mathbf T_j(r_g)^{-1}.
\end{equation}
The shared Gaussian transform is obtained by dual-quaternion
blending~\citep{kavan2007skinning}:
\begin{equation}
    \mathbf A_g(t)
    =
    \operatorname{DQB}\!\left(
        \left\{
        \overline a_{gj},
        \Delta\mathbf T_{gj}(t)
        \right\}_{j\in\mathcal N_g}
    \right).
\end{equation}
Because
$\Delta\mathbf T_{gj}(r_g)=\mathbf I$
for every $j$, the blended transformation also satisfies
$\mathbf A_g(r_g)=\mathbf I$.

\subsection{Local Hermite Motion Parameterization}
\label{sec:supp_local_motion}

Each Gaussian stores $L-1$ learnable six-dimensional controls.
We prepend a fixed zero control:
\begin{equation}
    \mathbf c_{g,0}=\mathbf 0,
    \qquad
    \mathbf c_{g,1},\ldots,
    \mathbf c_{g,L-1}\in\mathbb R^6.
\end{equation}
For normalized time $u\in[0,1]$, define
\begin{equation}
    k
    =
    \min\!\left(
        \lfloor(L-1)u\rfloor,
        L-2
    \right),
    \qquad
    s=(L-1)u-k.
\end{equation}
Let
\[
    k^-=\max(k-1,0),
    \qquad
    k^+=\min(k+2,L-1).
\]
The endpoint tangents for segment $k$ are computed from adjacent control
differences:
\begin{equation}
\begin{aligned}
    \mathbf m_{g,k}
    &=
    \frac{
        \mathbf c_{g,k+1}
        -
        \mathbf c_{g,k^-}
    }{
        k+1-k^-
    },
    \\
    \mathbf m_{g,k+1}
    &=
    \frac{
        \mathbf c_{g,k^+}
        -
        \mathbf c_{g,k}
    }{
        k^+-k
    }.
\end{aligned}
\label{eq:supp_local_tangents}
\end{equation}
Using the cubic Hermite bases
\begin{equation}
\begin{aligned}
    h_{00}(s)&=2s^3-3s^2+1,
    &
    h_{10}(s)&=s^3-2s^2+s,
    \\
    h_{01}(s)&=-2s^3+3s^2,
    &
    h_{11}(s)&=s^3-s^2,
\end{aligned}
\end{equation}
the local curve is
\begin{equation}
\begin{aligned}
    \mathbf h_g(u)
    &=
    h_{00}(s)\mathbf c_{g,k}
    +
    h_{10}(s)\mathbf m_{g,k}
    \\
    &\quad+
    h_{01}(s)\mathbf c_{g,k+1}
    +
    h_{11}(s)\mathbf m_{g,k+1}.
\end{aligned}
\label{eq:supp_local_hermite}
\end{equation}

For video time $t$, RayCo4D evaluates the difference
\begin{equation}
    \boldsymbol\delta_g(t)
    =
    \mathbf h_g
    \left(
        \frac{t}{\max(F-1,1)}
    \right)
    -
    \mathbf h_g
    \left(
        \frac{r_g}{\max(F-1,1)}
    \right).
\end{equation}
The subtraction makes the residual exactly zero at the Gaussian's own
reference time for every possible set of control values.
After scaling the translational part by the scaffold spatial unit
$\ell$, the local transform is
\begin{equation}
    \mathbf E_g(t)
    =
    \operatorname{Exp}\!\left(
        \left[
        \operatorname{diag}
        (\ell\mathbf I_3,\mathbf I_3)
        \boldsymbol\delta_g(t)
        \right]^\wedge
    \right).
\end{equation}

\subsection{Local Motion Regularization}
\label{sec:supp_local_regularization}

The weighted local-motion regularizer is
\begin{equation}
    \mathcal L_{\mathrm{local}}
    =
    \lambda_m\mathcal L_{\mathrm{mag}}
    +
    \lambda_c\mathcal L_{\mathrm{curv}}
    +
    \lambda_n\mathcal L_{\mathrm{nbr}}.
\label{eq:supp_local_total}
\end{equation}

\paragraph{Control magnitude.}
The magnitude term limits the overall capacity of the local correction:
\begin{equation}
    \mathcal L_{\mathrm{mag}}
    =
    \frac{1}{6G(L-1)}
    \sum_{g=1}^{G}
    \sum_{k=1}^{L-1}
    \|\mathbf c_{g,k}\|_2^2.
\label{eq:supp_local_magnitude}
\end{equation}

\paragraph{Control curvature.}
Temporal smoothness is encouraged through second differences of the
control sequence:
\begin{equation}
    \mathcal L_{\mathrm{curv}}
    =
    \frac{1}{6G(L-2)}
    \sum_{g=1}^{G}
    \sum_{k=1}^{L-2}
    \|
        \mathbf c_{g,k+1}
        -
        2\mathbf c_{g,k}
        +
        \mathbf c_{g,k-1}
    \|_2^2.
\label{eq:supp_local_curvature}
\end{equation}
For $L=2$, this term is set to zero.
It acts as a smoothness prior on the control sequence.

\paragraph{Neighborhood consistency.}
Local-motion neighborhoods are constructed in a common scene
configuration.
We first transport every Gaussian to
\[
    t_\star=\lfloor F/2\rfloor
\]
using the shared motion only:
\begin{equation}
    \mathbf P_g^0(t)
    =
    \mathbf A_g(t)\mathbf P_g^{\mathrm{ref}}.
\end{equation}
For each Gaussian, we retain up to five nearest neighbors in this common
configuration.
The edge weight between Gaussians $g$ and $h$ is
\begin{equation}
    \beta_{gh}
    =
    \exp\!\left(
        -
        \frac{
        \|
        \boldsymbol\mu_g^0(t_\star)
        -
        \boldsymbol\mu_h^0(t_\star)
        \|_2^2
        }{
        2\ell^2
        }
    \right)
    \frac{
        |\mathcal N_g\cap\mathcal N_h|
    }{
        K
    }.
\label{eq:supp_local_edge_weight}
\end{equation}
The first factor favors spatially close Gaussians, and the second factor
measures the overlap of their shared-node support.

For an adjacent-frame interval $[a,b]$, define the local control
increment
\begin{equation}
    \Delta\mathbf h_g^{a,b}
    =
    \mathbf h_g
    \left(
        \frac{b}{\max(F-1,1)}
    \right)
    -
    \mathbf h_g
    \left(
        \frac{a}{\max(F-1,1)}
    \right).
\end{equation}
Let $\mathbf R_g^0(t)$ be the Gaussian orientation produced by shared
motion at the sampled scene time $t$.
We express the increment in a common world orientation as
\begin{equation}
    \mathbf d_g(t)
    =
    \operatorname{diag}
    \left(
        \mathbf R_g^0(t),
        \mathbf R_g^0(t)
    \right)
    \Delta\mathbf h_g^{a,b}.
\label{eq:supp_oriented_increment}
\end{equation}
The neighborhood term is
\begin{equation}
    \mathcal L_{\mathrm{nbr}}
    =
    \frac{
        \sum_{(g,h)}
        \beta_{gh}
        \|
        \mathbf d_g(t)-\mathbf d_h(t)
        \|_2^2
    }{
        6
        \sum_{(g,h)}
        \beta_{gh}
    }.
\label{eq:supp_neighbor_regularization}
\end{equation}
In implementation, this objective is estimated from sampled center
Gaussians and their cached neighbors.

\subsection{Dynamic Gaussian Parameters}
\label{sec:supp_dynamic_rendering}

Write the composed Gaussian pose as
\[
    \mathbf P_g(t)
    =
    [\mathbf R_g(t),\boldsymbol\mu_g(t)].
\]
For a learned time-independent scale vector
$\mathbf s_g\in\mathbb R_{>0}^3$,
the world-space covariance is
\begin{equation}
    \boldsymbol\Sigma_g(t)
    =
    \mathbf R_g(t)
    \operatorname{diag}(\mathbf s_g^2)
    \mathbf R_g(t)^\top.
\label{eq:supp_gaussian_covariance}
\end{equation}
The center and covariance vary through the hierarchical motion, while
scale, opacity, and spherical-harmonic coefficients are learned as
Gaussian attributes shared across scene times.
The dynamic and static Gaussian sets are concatenated before
differentiable rasterization.

\subsection{Training Objective and Auxiliary Terms}
\label{sec:supp_training_objective}

The main objective is reproduced here for completeness:
\begin{equation}
\begin{aligned}
    \mathcal L
    &=
    \mathcal L_{\mathrm{scene}}
    +\lambda_u\mathcal L_{\mathrm{2D}}
    +\lambda_d\mathcal L_{\mathrm{depth}}
    +\lambda_W\mathcal L_{\mathrm{WNOA}}
    +\lambda_A\mathcal L_{\mathrm{ARAP}}
    \\
    &\quad
    +\mathcal L_{\mathrm{local}}
    +\mathcal L_{\mathrm{cal}}
    +\lambda_{\mathrm{warp}}
     \mathcal L_{\mathrm{warp}}.
\end{aligned}
\end{equation}
The scene objective combines RGB reconstruction, rendered depth,
rendered track correspondence, coverage, ownership, and binding-offset
regularization. RGB supervision uses pixel-averaged $\ell_1$ error plus
$0.1(1-\mathrm{SSIM})$. The RGB, rendered-depth, correspondence, and
binding-offset terms have weights $1$, $0.01$, $0.005$, and $0.01$,
respectively; track coverage, ownership coverage, and ownership
classification each use $0.1$. Ownership classification is linearly
annealed to zero during joint fitting.

\paragraph{Adjacent-frame appearance transport.}
For adjacent frames $f$ and $f'$, the predicted Gaussian poses define a
local image transport
$\mathcal W_{f\rightarrow f'}$.
The target image is sampled at the transported coordinates, yielding
the appearance consistency term
\begin{equation}
    \mathcal L_{\mathrm{warp}}
    =
    \operatorname{LBMean}
    \left(
        \left\|
        I_{f'}
        \bigl(
        \mathcal W_{f\rightarrow f'}(\mathbf p)
        \bigr)
        -
        I_f(\mathbf p)
        \right\|_1,
        \omega(\mathbf p)
    \right),
\label{eq:supp_warp_loss}
\end{equation}
where $\operatorname{LBMean}$ denotes layer-balanced averaging and
$\omega$ combines static/dynamic ownership with externally estimated
correspondence-validity support.
Predicted alpha and depth determine image coverage and occlusion at the
transported location.
A companion coverage term encourages sufficient Gaussian support at
valid transported pixels.

\paragraph{Camera calibration.}
When camera refinement is enabled, the camera-to-world pose is written as
\begin{equation}
    \mathbf C_f
    =
    \operatorname{Exp}
    (\boldsymbol\epsilon_f^\wedge)
    \mathbf C_f^0,
    \qquad
    \boldsymbol\epsilon_0=\mathbf 0,
\label{eq:supp_camera_correction}
\end{equation}
where $\mathbf C_f^0$ is the input pose.
Intrinsics remain fixed.
The camera regularizers are
\begin{equation}
\begin{aligned}
    \mathcal L_{\mathrm{cam}}
    &=
    \frac{
        \sum_{f=1}^{F-1}
        \|\boldsymbol\epsilon_f\|_2^2
    }{
        6(F-1)
    },
    \\
    \mathcal L_{\mathrm{cam\text{-}smooth}}
    &=
    \frac{
        \sum_{f=0}^{F-2}
        \|
        \boldsymbol\epsilon_{f+1}
        -
        \boldsymbol\epsilon_f
        \|_2^2
    }{
        6(F-1)
    }.
\end{aligned}
\label{eq:supp_camera_regularization}
\end{equation}

\paragraph{Depth calibration.}
We optimize a scalar log-depth correction $s_f$ per frame, centered
across the sequence:
\begin{equation}
\begin{aligned}
    \kappa_f &= s_f-\frac{1}{F}\sum_{h=0}^{F-1}s_h, \\
    D_f^{\mathrm{corr}}(\mathbf u)
    &=D_f(\mathbf u)\exp(\kappa_f).
\end{aligned}
\label{eq:supp_depth_calibration}
\end{equation}
The optional spatial correction grid is fixed to zero.
The calibration objective combines penalties on these corrections with
the camera regularizers and static-track constraints:
\begin{equation}
    \mathcal L_{\mathrm{cal}}
    =
    \lambda_{\mathrm{cam}}
    \mathcal L_{\mathrm{cam}}
    +
    \lambda_{\mathrm{cam\text{-}smooth}}
    \mathcal L_{\mathrm{cam\text{-}smooth}}
    +
    \lambda_{\mathrm{dep\text{-}cal}}
    \mathcal L_{\mathrm{dep\text{-}cal}}
    +
    \lambda_{\mathrm{static}}
    \mathcal L_{\mathrm{static}}.
\end{equation}

Here, $\mathcal L_{\mathrm{dep\text{-}cal}}$ penalizes squared
log-depth corrections at valid depth observations, weighted by depth
reliability; $\mathcal L_{\mathrm{static}}$ is the confidence-weighted
smooth-$\ell_1$ reprojection error of fixed static landmarks and updates
only the camera corrections.

\paragraph{Loss weights.}
We set $(\lambda_u,\lambda_d,\lambda_W,\lambda_A)=(1,1,0.1,0.1)$ and
$(\lambda_m,\lambda_c,\lambda_n)=(0.001,0.01,0.01)$.
The camera prior, camera smoothness, depth-calibration prior, and static
landmark weights are $0.001$, $0.0001$, $0.1$, and $0.01$, respectively.
The depth-calibration prior is evaluated on both sampled ray observations
and the rendered source frame. Appearance transport and its companion
coverage term use weights $0.05$ and $0.1$.

\Needspace{22\baselineskip}
\suppressfloats[t]
\subsection{Optimization Schedule}
\label{sec:supp_optimization_schedule}

\begingroup
\setlength{\intextsep}{0pt}
\begin{wraptable}{r}{0.58\textwidth}
\centering
\setlength{\abovecaptionskip}{0pt}
\caption{Training stages. Iterations are counted separately for each row.}
\label{tab:optimization_schedule}
\setlength{\tabcolsep}{3pt}
\renewcommand{\arraystretch}{1.05}
\begin{tabular*}{\linewidth}{@{\extracolsep{\fill}}p{0.25\linewidth}rp{0.50\linewidth}@{}}
\toprule
Stage & Iterations & \raggedright Optimized variables \tabularnewline
\midrule
\raggedright Motion warm-up & 4{,}000 & \raggedright Node states and query depths \tabularnewline
\raggedright Gaussian warm-up & 1{,}000 & \raggedright Gaussian geometry and appearance \tabularnewline
\raggedright Joint fitting & 6{,}000 & \raggedright Scene, motion, and calibration \tabularnewline
\raggedright Final fitting & 1{,}000 & \raggedright Continuous variables; fixed structure \tabularnewline
\bottomrule
\end{tabular*}
\end{wraptable}
Tab.~\ref{tab:optimization_schedule} summarizes the four stages.
After motion warm-up, step indices below refer to the separate
8{,}000-step scene-fitting stage, starting at zero.
Gaussian binding offsets, camera--depth calibration, local Hermite
controls, and appearance transport are enabled at steps 1{,}000,
2{,}000, 2{,}500, and 3{,}000, respectively.
Gaussian parameters are updated every step; node states, query depths,
and active calibration variables are updated every five steps after
Gaussian warm-up, including the final stage.
Gaussian structure control is active at steps 1{,}000--5{,}999;
node adaptation is restricted to joint fitting.
Ownership is initialized before Gaussian fitting, corrected once at
step 3{,}000, and then fixed.
\par
\endgroup

\subsection{Additional Implementation Details}
\label{sec:addition_implementation_details}

\paragraph{Initialization and representation.}
Lifted tracks initialize motion nodes through spatial coverage sampling;
the node count depends on the scene.
Missing positions are interpolated linearly between observations and
extended from the nearest endpoint outside the observed span.
These filled positions are used only for initialization and are not
marked as measured support.
Pose--velocity supports are placed every two frames, including both
endpoints; initial rotations are identities and body velocities are
estimated from neighboring support poses.
We use $K=16$ candidate nodes, scaffold scale $\ell=0.02$ in normalized
scene units, $q_p=q_R=1$, and $L=8$ local Hermite controls.
Initialization uses at most 80{,}000 static depth samples, 80{,}000 dynamic
track seeds, and 40{,}000 additional dynamic depth samples.

\paragraph{Sampling and updates.}
Each motion update samples up to 16{,}384 visible observations with
replacement. Motion regularization uses a contiguous window of up to
16 support states and up to 1{,}024 sampled nodes; this is a minibatch
size, not the total node count.
Track candidates are refreshed every 100 motion-warm-up steps.
During scene fitting, binding refreshes occur at initialization,
step 3{,}000, and as required by node-structure changes.
Layer supervision is evaluated every two steps and appearance transport
every four steps, with their losses scaled by the respective intervals.

\paragraph{Optimization.}
We use Adam with the learning rates in Tab.~\ref{tab:optimization_settings}.
During motion warm-up, node and query-depth rates follow a cosine decay
to $5\%$ of their initial values; scene fitting starts these groups at
$10\%$ of their initial warm-up rates.
From scene-fitting step 2{,}000, Gaussian position rates decay
exponentially to $1.6\times10^{-6}$, and camera and depth-scale rates
to one tenth of their initial values. The other rates remain constant.

\begin{table}[t]
\centering
\caption{Initial learning rates. Node and query-depth entries refer to
motion warm-up; their scene-fitting rates are scaled by $0.1$.}
\label{tab:optimization_settings}
\begin{tabular}{lr}
\toprule
Parameter group & Learning rate \\
\midrule
Reference-query depths & $10^{-2}$ \\
Node translation, rotation, and body velocity & $3\times10^{-3}$ \\
Gaussian position & $1.6\times10^{-4}$ \\
Gaussian rotation / scale / opacity & $10^{-3}$ / $5\times10^{-3}$ / $5\times10^{-2}$ \\
SH coefficients: constant / higher-order & $2.5\times10^{-3}$ / $1.25\times10^{-4}$ \\
Gaussian binding offsets & $10^{-2}$ \\
Local Hermite controls & $10^{-3}$ \\
Camera translation and rotation corrections & $3\times10^{-5}$ \\
Per-frame log-depth scales & $10^{-3}$ \\
\bottomrule
\end{tabular}
\end{table}

\subsection{Baseline Details}
\label{sec:baseline_details}

We compare against methods covering the main motion representations used
in recent monocular dynamic reconstruction.
MoSca~\citep{lei2025mosca} constructs a sparse motion scaffold from lifted
tracks and propagates node transformations to dynamic Gaussians.
Shape of Motion~\citep{wang2025shapeofmotion} represents scene dynamics
using shared $\mathrm{SE}(3)$ motion bases constrained by long-range tracks
and monocular geometry.
SE(3) B-splineGS~\citep{zhang2026se3bspline} parameterizes continuous
motion using $\mathrm{SE}(3)$ B-spline control trajectories, providing the
most direct comparison to our WNOA pose--velocity formulation.
OriGS~\citep{wu2025origs} and RiGS~\citep{wu2026rigs} are included for
their treatment of complex local deformation and multi-timescale motion,
respectively.
USPLAT4D~\citep{guo2026usplat4d} evaluates reconstruction under incomplete
and unreliable observations, including occlusion.

\subsection{Detail Datasets and Evaluation Protocols}
\label{sec:dataset_details}
\paragraph{DyCheck Dataset.}
The DyCheck Dataset~\citep{gao2022dynamic} contains casually
captured monocular videos of dynamic scenes together with spatially
separated evaluation views.
It is particularly challenging because the training and evaluation
cameras often observe only partially overlapping scene regions.
We follow the standard seven-scene protocol at $2\times$ resolution.
Novel-view synthesis metrics are computed within the provided
co-visibility masks, yielding mPSNR, mSSIM, and mLPIPS.

We evaluate motion correspondence on DyCheck using the percentage of
correctly transferred keypoints (PCK-T), following the DyCheck
protocol~\citep{gao2022dynamic} as in
OriGS~\citep{wu2025origs} and SE(3) B-splineGS~\citep{zhang2026se3bspline}.
All methods use the same query points, target frames, visibility masks,
and distance threshold. RayCo4D propagates reference queries through
its shared motion field and projects them into the target camera;
only observations visible in both frames are evaluated.

\paragraph{NVIDIA Dynamic Scene Dataset.}
The NVIDIA Dynamic Scene Dataset~\citep{yoon2020novel} consists of
synchronized multi-camera recordings of dynamic scenes.
Following prior monocular dynamic reconstruction methods, we construct
each training sequence from a single moving-camera video and reserve the
remaining synchronized views for evaluation.
We report PSNR, SSIM, and LPIPS between the rendered and ground-truth
test images using the standard image resolution and evaluation split.

\paragraph{DAVIS.}
DAVIS~\citep{perazzi2016davis} contains in-the-wild monocular videos with
challenging non-rigid motion, occlusion, motion blur, and complex object
interactions.
We use DAVIS to qualitatively assess reconstruction stability on scenes
outside controlled multi-view capture settings.
Because DAVIS does not provide synchronized images from held-out cameras,
we do not report quantitative novel-view synthesis results on this
dataset.

\paragraph{Continuous-time evaluation.}
Selected intermediate frames are excluded from image, track, and depth
supervision. All compared methods use the same observed frames and
held-out timestamps. We render the corresponding camera views and report
PSNR, SSIM, and LPIPS on NVIDIA (Tab.~\ref{tab:rayco4d_intermediate});
Fig.~\ref{fig:supp_continuous_time} additionally shows qualitative results
on DAVIS and DyCheck.

\suppressfloats[t]
\subsection{Runtime Comparison}
\label{sec:supp_runtime}

Tab.~\ref{tab:supp_runtime} reports average wall-clock times for 200-frame sequences from DyCheck at the $2\times$ resolution setting. Pre. denotes preprocessing, and Train. denotes reconstruction training. For RayCo4D, training includes motion-scaffold optimization and the three scene-fitting stages in Tab.~\ref{tab:optimization_schedule}.

\begingroup
\setlength{\intextsep}{0pt}
\Needspace{13\baselineskip}
\suppressfloats[t]
\begin{wraptable}{r}{0.50\textwidth}
\centering
\setlength{\abovecaptionskip}{0pt}
\caption{\textbf{Average runtime on DyCheck (min).}}
\label{tab:supp_runtime}
\normalsize
\setlength{\tabcolsep}{1pt}
\renewcommand{\arraystretch}{1.05}
\begin{tabular*}{\linewidth}{@{}l@{\extracolsep{\fill}\hspace{2pt}}r@{\extracolsep{0pt}\hspace{3pt}}r@{}}
\toprule
\textbf{Method} & Pre. & Train. \\
\midrule
MoSca~{\footnotesize \citep{lei2025mosca}} & 20 & 40 \\
Shape of Motion~{\footnotesize \citep{wang2025shapeofmotion}} & 35 & 63 \\
OriGS~{\footnotesize \citep{wu2025origs}} & 30 & 77 \\
RiGS~{\footnotesize \citep{wu2026rigs}} & 37 & 56 \\
SE(3) B-splineGS~{\footnotesize \citep{zhang2026se3bspline}} & 27 & 61 \\
USPLAT4D~{\footnotesize \citep{guo2026usplat4d}} & 60 & 20 \\
\midrule
\textbf{RayCo4D} & 36 & 50 \\
\bottomrule
\end{tabular*}
\end{wraptable}
USPLAT4D~\citep{guo2026usplat4d} uses MoSca-generated Gaussian priors~\citep{lei2025mosca}; its 60-minute Pre. entry includes the cost of generating this initialization.

RayCo4D takes 86 minutes in total, compared with 88--107 minutes for Shape of Motion, OriGS, RiGS, and SE(3) B-splineGS. Its 50-minute training stage is also shorter than the 56--77 minutes required by these four methods. MoSca has a lower total runtime of 60 minutes. USPLAT4D takes 80 minutes in total, including 60 minutes for preprocessing and 20 minutes for training.\csname @ifpackageloaded\endcsname{RayCo4D}{\space}{}Together with the DyCheck results in Tab.~\ref{tab:main_results}, these timings characterize the trade-off between reconstruction quality and computational cost.
\par
\endgroup

\section{Additional Experimental Results}
\label{sec:supp_additional_results}

\subsection{Continuous-Time Reconstruction}
\label{sec:supp_continuous_qualitative}
\begin{figure}[H]
    \centering
    \includegraphics[width=\linewidth]{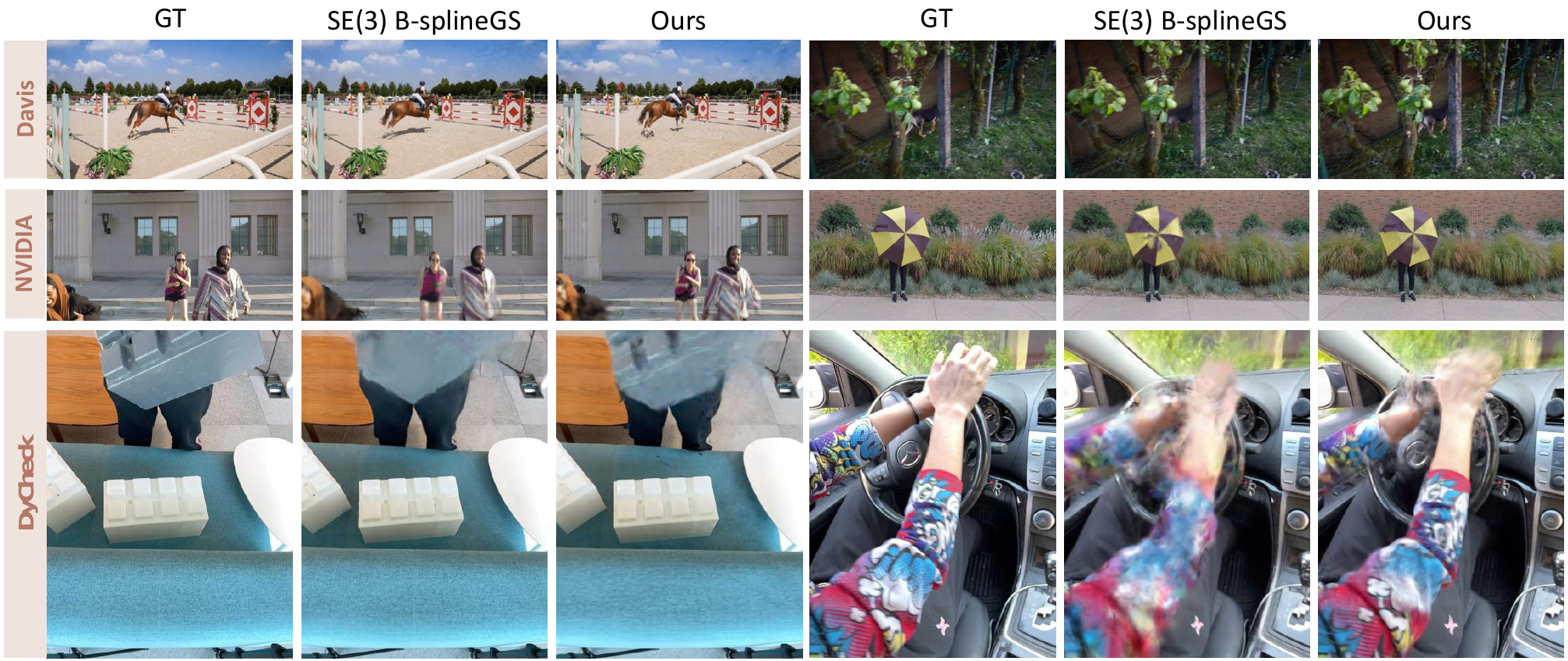}
    \caption{
        Continuous-time reconstruction on DAVIS, NVIDIA, and DyCheck.
        GT denotes intermediate frames excluded from image, track,
        and depth supervision. All renderings match the GT timestamps
        and camera views.
    }
    \label{fig:supp_continuous_time}
\end{figure}

Fig.~\ref{fig:supp_continuous_time} complements the NVIDIA results in
Tab.~\ref{tab:rayco4d_intermediate} with qualitative comparisons on
DAVIS, NVIDIA, and DyCheck, following Sec.~\ref{sec:dataset_details}.

\subsection{Challenging Cases and Limitations}
\label{sec:supp_challenging_cases}
Long intervals without reliable observations leave the intervening motion
ambiguous. If motion changes abruptly within such an interval, the locally
smooth WNOA prior may yield inaccurate intermediate motion.

\begin{figure}[H]
    \centering
    \includegraphics[width=\linewidth]{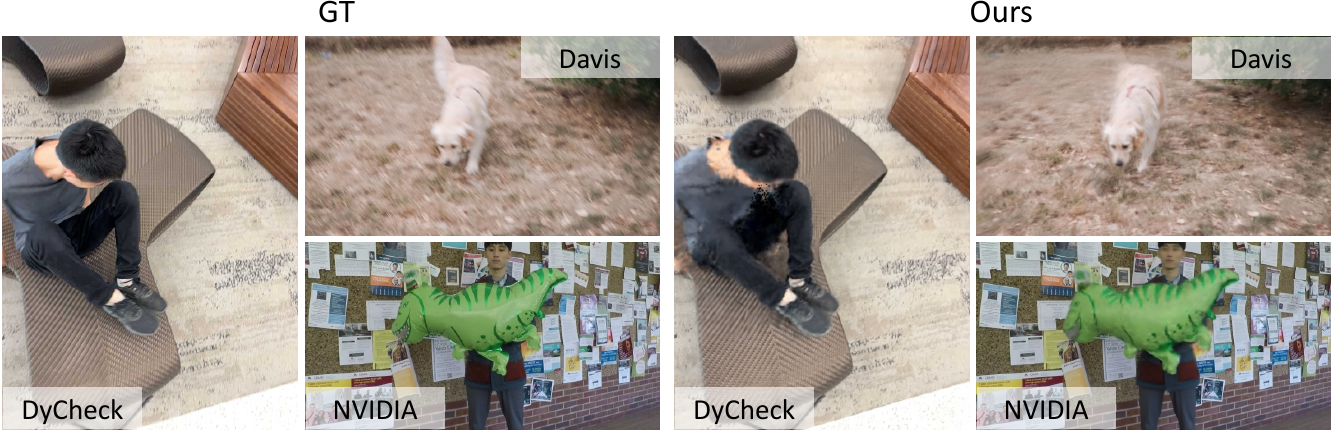}
    \caption{
        Challenging reconstruction examples on DyCheck, DAVIS, and
        NVIDIA. GT denotes held-out real frames excluded from training;
        RayCo4D renderings use the corresponding timestamps and camera views.
    }
    \label{fig:supp_failure_cases}
\end{figure}

The measurement--state formulation separates motion-node identity from
individual tracks, but the latent states still rely on geometric observations.
Large errors in tracks, estimated camera poses, or monocular depths can
therefore distort the recovered motion and geometry. Future work could model
measurement uncertainty, adapt temporal support to motion changes, and use
stronger scene-level priors for poorly observed intervals.
Fig.~\ref{fig:supp_failure_cases} shows examples of remaining
reconstruction errors.

\section{Additional Qualitative Comparisons}
\label{sec:supp_additional_qualitative}
\begin{figure}[H]\centering\setlength{\abovecaptionskip}{4pt}\setlength{\belowcaptionskip}{0pt}
    \includegraphics[width=\linewidth]{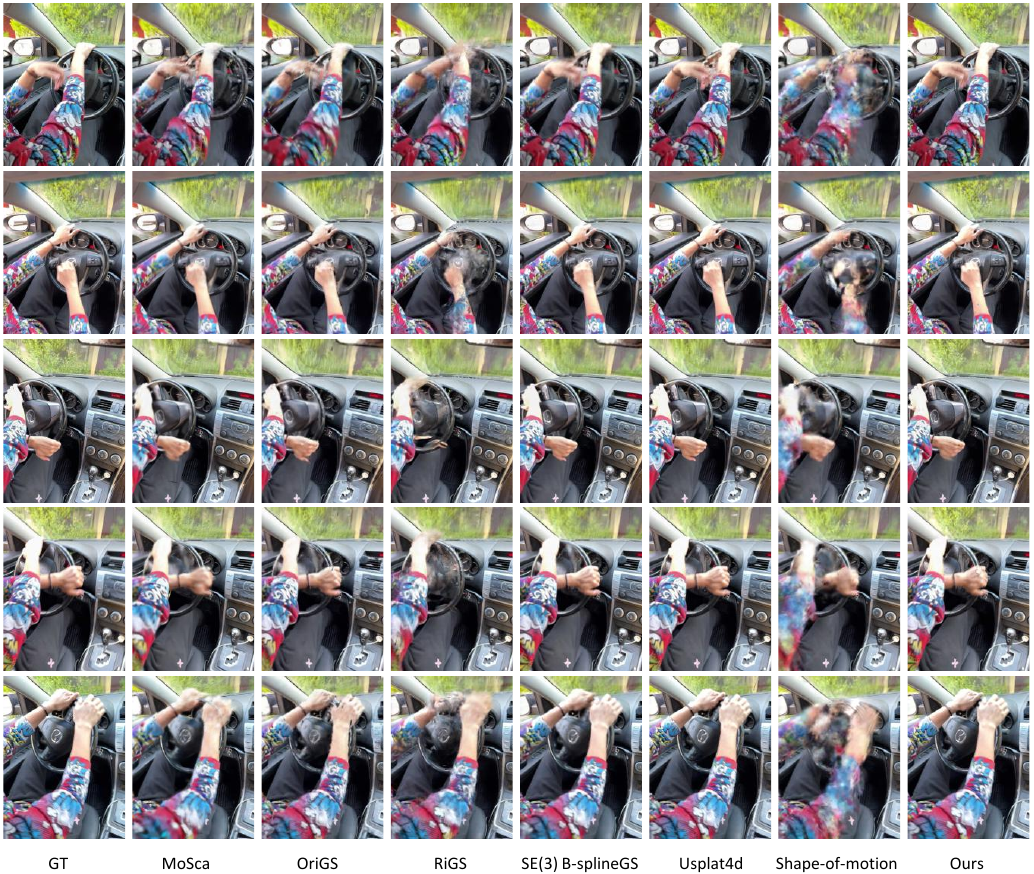}
    \caption{Visual comparison of novel-view synthesis on the ``Wheel'' scene of the DyCheck dataset~\citep{gao2022dynamic}.}
    \label{fig:supp_dycheck_paper_windmill}\medskip
    \includegraphics[width=\linewidth]{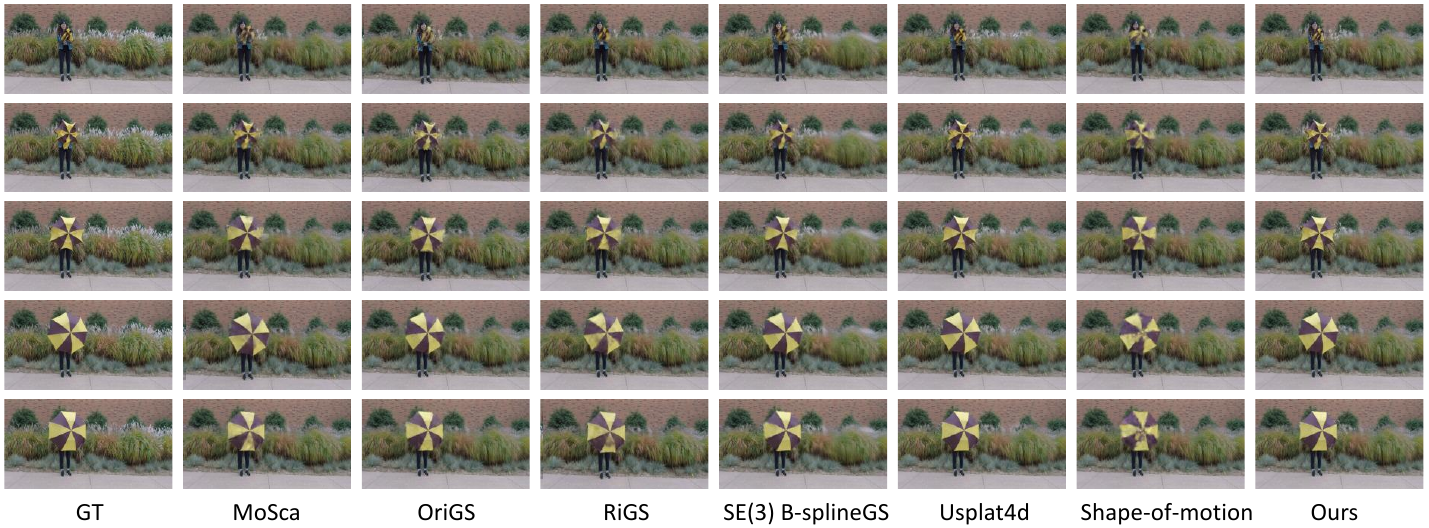}
    \caption{Visual comparison of novel-view synthesis on the ``Umbrella'' scene of the NVIDIA dataset~\citep{yoon2020novel}.}
    \label{fig:supp_nvidia_umbrella}\end{figure}

\begin{figure}[H]\centering\setlength{\abovecaptionskip}{4pt}\setlength{\belowcaptionskip}{0pt}
    \includegraphics[width=\linewidth]{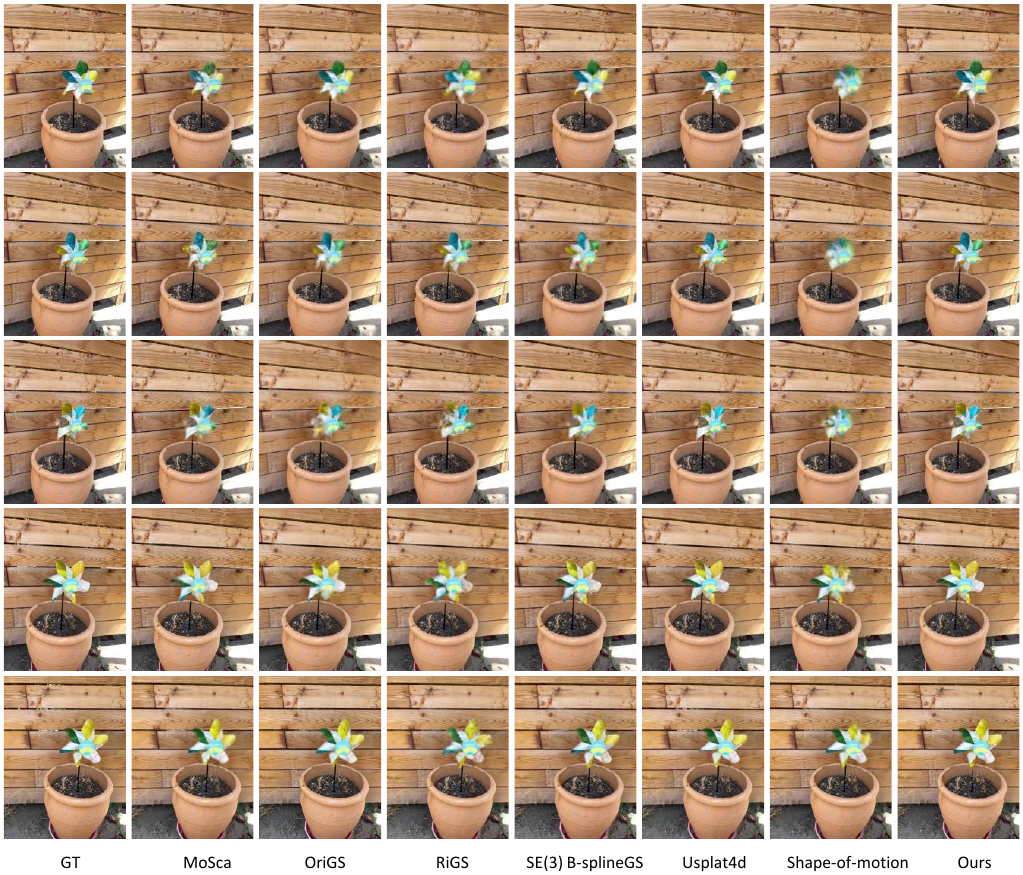}
    \caption{Visual comparison of novel-view synthesis on the ``Paper-windmill'' scene of the DyCheck dataset~\citep{gao2022dynamic}.}
    \label{fig:supp_dycheck_wheel_clean}
\end{figure}

Figs.~\ref{fig:supp_davis_breakdance}--\ref{fig:supp_dycheck_wheel_clean}
provide scene-wise qualitative comparisons of the full RayCo4D model
and the baselines on DAVIS, NVIDIA, and DyCheck,
supplementing Fig.~\ref{fig:qualitative} in the main paper.


\begingroup
\setlength{\intextsep}{4pt}
\begin{figure}[H]
    \centering
    \setlength{\abovecaptionskip}{4pt}
    \setlength{\belowcaptionskip}{0pt}
    \includegraphics[width=0.94\linewidth]{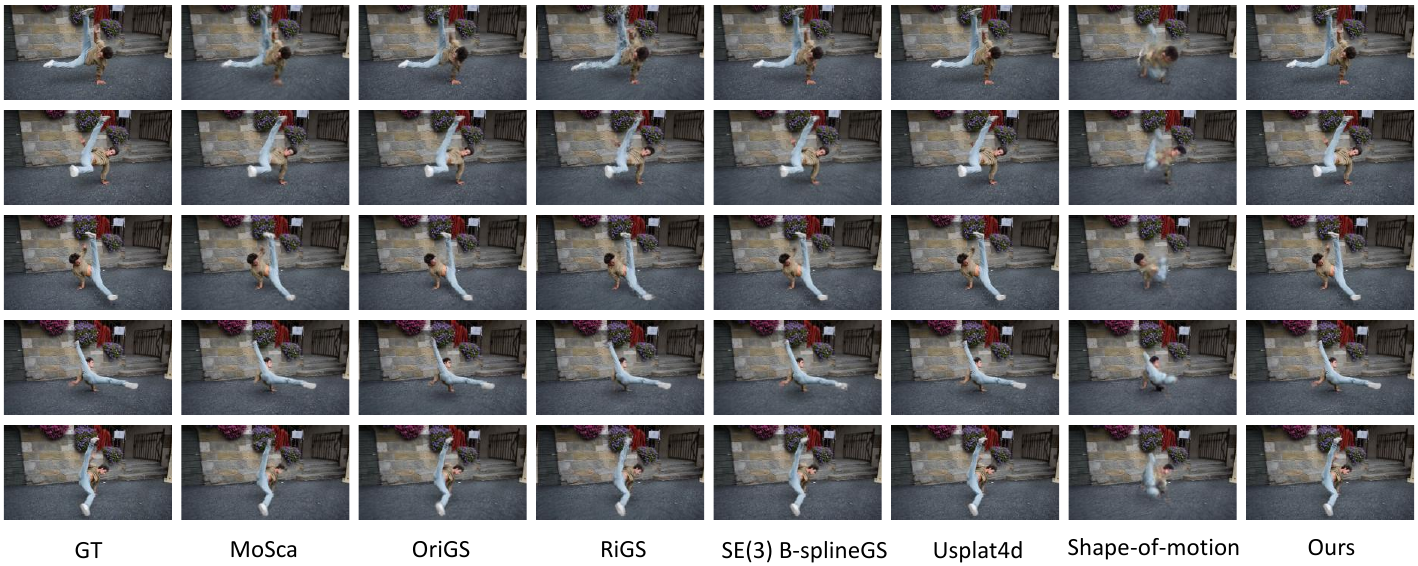}
    \caption{Qualitative reconstruction comparison on the ``Breakdance-flare'' scene of DAVIS~\citep{perazzi2016davis}.}
    \label{fig:supp_davis_breakdance}
    \medskip
    \includegraphics[width=0.94\linewidth]{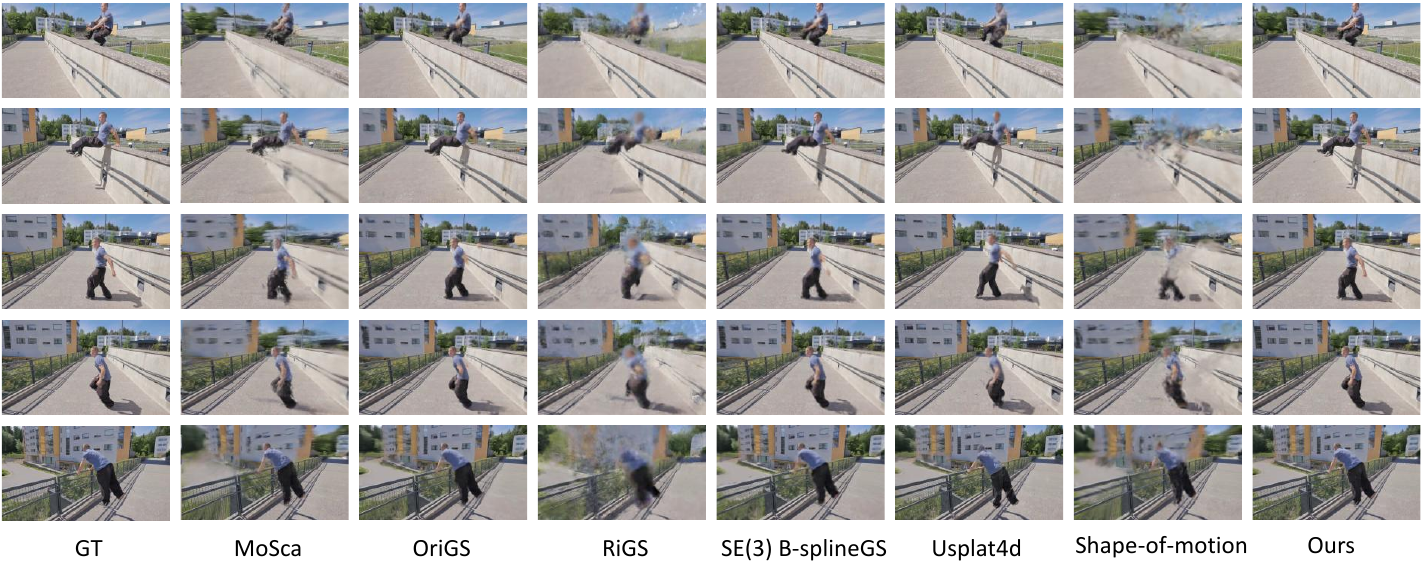}
    \caption{Qualitative reconstruction comparison on the ``Parkour'' scene of DAVIS~\citep{perazzi2016davis}.}
    \label{fig:supp_davis_parkour}
\medskip
    \includegraphics[width=0.94\linewidth]{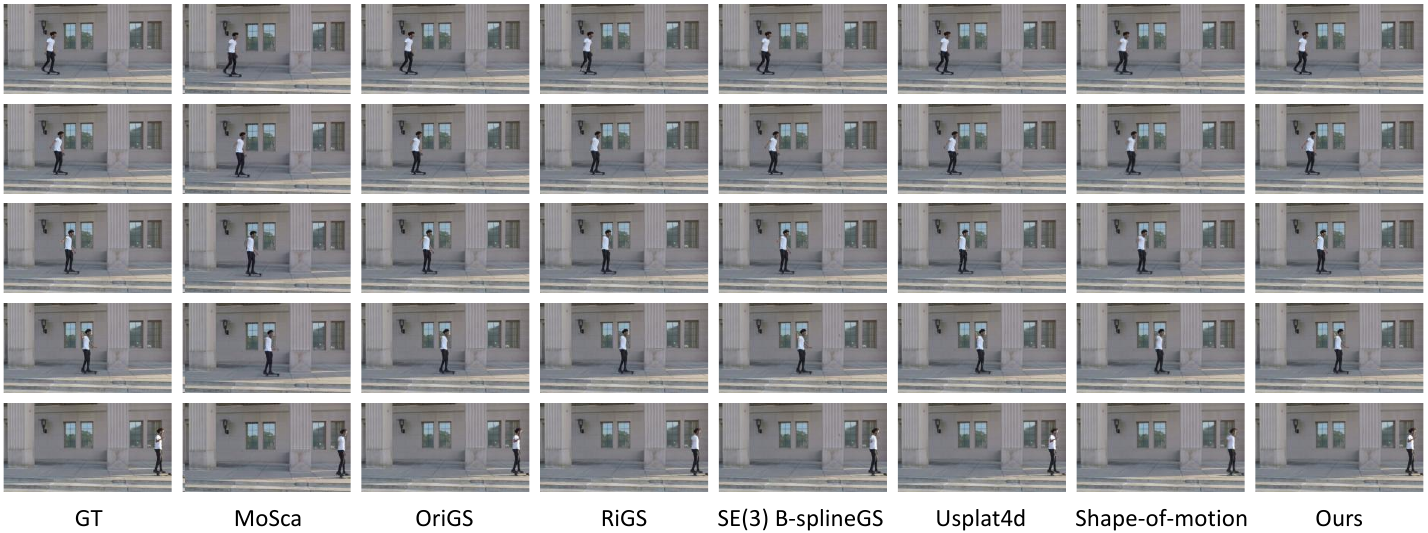}
    \caption{Visual comparison of novel-view synthesis on the ``Skating'' scene of the NVIDIA dataset~\citep{yoon2020novel}.}
    \label{fig:supp_nvidia_skating}\end{figure}\endgroup\clearpage\clearpage

\end{document}